%% file: main_for_arxiv.tex
\documentclass[journal]{IEEEtran}
\ifCLASSINFOpdf
\else
\fi
\input{macros}

\usepackage{array}
\usepackage{multirow}
\usepackage{enumerate}
\usepackage{makecell}
\usepackage{tabularx} 
\usepackage{algorithm,algcompatible}
\usepackage{lipsum}
\usepackage{bbm}
\usepackage{times}
\usepackage[cmex10]{amsmath}
\usepackage{algpseudocode}
\usepackage{epsfig,verbatim}
\usepackage{amssymb} 
\usepackage{graphics,subfigure}
\usepackage{xcolor}
\usepackage{graphicx}
\usepackage{tabularx,booktabs,hhline}
\usepackage{color, colortbl}
\definecolor{LightCyan}{rgb}{0.88,1,1}
\definecolor{lightgray}{gray}{0.95}
\usepackage{multirow}

\newtheorem{theorem}{Theorem}

\newtheorem{lemma}{Lemma}

\newtheorem{remark}{Remark}

\newcommand{\argmin}{\operatornamewithlimits{argmin}}

\begin{document}
%
\title{Distributed Online Learning with Multiple Kernels}

%
%
%

\author{Jeongmin Chae,~\IEEEmembership{Student,~IEEE,}
        and~Songnam Hong,~\IEEEmembership{Member,~IEEE}
\thanks{J. Chae is with the Department of Electrical Engineering, University of Southern California, CA, 90089, USA (e-mail: chaej@usc.edu)}
\thanks{S. Hong is with the Department of Electronic Engineering, Hanyang University, Seoul, 04763, Korea (e-mail: snhong@hanyang.ac.kr)} 

}

\maketitle

\begin{abstract} 

In the Internet-of-Things (IoT) systems, there are plenty of informative data provided by a massive number of IoT devices (e.g., sensors). Learning a function from such data is of great interest in machine learning tasks for IoT systems. Focusing on streaming (or sequential) data, we present a privacy-preserving distributed online learning framework with multiple kernels (named DOMKL). The proposed DOMKL is devised by leveraging the principles of an online alternating direction of multipliers (OADMM) and a distributed Hedge algorithm. We theoretically prove that  DOMKL over $T$ time slots can achieve an optimal sublinear regret $\Oc(\sqrt{T})$, implying that every learned function achieves the performance of the best function in hindsight as in the state-of-the-art centralized online learning method. Moreover, it is ensured that the learned functions of any two neighboring learners have a negligible difference as $T$ grows, i.e., the so-called consensus constraints hold. Via experimental tests with various real datasets, we verify the effectiveness of the proposed DOMKL on regression and time-series prediction tasks.

\end{abstract}


\IEEEpeerreviewmaketitle

%
%
\section{Introduction}\label{sec:intro}




The Internet-of-Things (IoT) systems consist of a massive number of machine-type devices (e.g., sensors) which can be employed to monitor and analyze various cyber-physical systems such as smart cities, smart grids, connected cars, mobile healthcare, intelligent energy management, traffic management, and so on \cite{zanella2014internet,shah2016survey,bedi2018review}. In order to accomplish sophisticated tasks, the IoT systems are expected to sense, learn, and adapt to sequentially arriving measurements. These can be efficiently performed via machine learning techniques \cite{siryani2017machine, li2018learning, liang2018toward}, where a large and complex data can be efficiently analyzed in a timely fashion. In particular, this paper focuses on learning a function as it is of great interest in various machine learning tasks such as classification, regression, clustering, dimensionality reduction, and reinforcement learning \cite{shawe2004kernel, lin2010multiple, chen2013online, vanli2016sequential}. In the most of existing works,  function learning tasks are performed in a centralized fashion with the assumption that all the data (possibly acquired by distributed IoT devices) are gathered in the cloud (or central processor). This assumption, however, is not applicable for the IoT systems due to the limitations of communication bandwidth, power consumption, latency, and privacy concern. Also, data privacy has become a growing concern in IoT systems. Multiple cases of data leakage and misuse have demonstrated that the centralized processing of learning a function comes at high risk for the end users privacy. As IoT devices usually collect data in privacy environments, these concerns hold particularly strong. Thus, a privacy-preserving distributed function learning, where training data is stored locally in every distributed device (or learner) without revealing the local data to a cloud, has been investigated \cite{lin2010design,chen2014consensus,gursoy2016privacy, sattler2019robust}. Moreover, in many real-world applications, function learning tasks are expected to be performed in an online fashion. For instance, online learning is required when data arrive sequentially such as solar power prediction \cite{kraemer2020operationalizing}, time series prediction \cite{richard2008online}, energy-saving in smart home \cite{zhou2019privacy}, and  when the large number of data makes it impossible to carry out data analytic in batch form \cite{kivinen2004online}. Motivated by this, we investigate a distributed function learning framework suitable for streaming (or sequential) data.


Supervised function learning tasks, which are closely related to the subject of this paper, are formulated as follows.  Given the training data $\{(\xv_t,y_t): t=1,...,T\}$ with features $\xv_t \in \RR^d$ and labels $y_t \in \RR$, the primary goal of these tasks is to learn (or estimate) a function $f:\RR^d\rightarrow \RR$ which minimizes the accumulate loss $\frac{1}{T}\sum_{t=1}^T\Lc(f(\xv_t),y_t)$,
where $f(\xv_t)=\hat{y}_t$ and  $\Lc(\cdot,\cdot)$ represent an estimated label and a loss function, respectively. This challenging problem can be tractable with the restriction that $f(\cdot)$ belongs to a well-structured function space such as reproducing kernel Hilbert space (RKHS) \cite{scholkopf2001learning}. Such function learning is called kernel-based learning. Definitely the accuracy of a kernel-based learning fully relies on a preselected kernel, which is determined manually either by a task-specific priori knowledge or by some intensive cross-validation process. 
Multiple kernel learning (MKL), using a predetermined set of kernels (called a kernel dictionary), is more powerful as it can enable a data-driven kernel selection from a given dictionary. That is, a linear (or non-linear) combination of multiple kernels is optimized as the consequence of a function learning process \cite{gonen2011multiple}.




Recently, {\em online} MKL (OMKL) has been proposed, which seeks the optimal combination of multiple kernel functions in an online fashion. Two popular approaches to learn the best kernel combination are the Hedge algorithm and online gradient descent (OGD) algorithm \cite{sahoo2014online, bubeck2011introduction}. It was shown in  \cite{kivinen2004online,sahoo2014online,shen2019random,hong2020active} that OMKL can yield an attractive accuracy performance and enjoy a great flexibility compared with single-kernel online learning. Whereas, OMKL generally suffers from a high computational complexity as the dimension of optimization variables grow with time (i.e., the number of data $T$) \cite{shawe2004kernel,wahba1990spline}. Recently in \cite{shen2019random}, this problem has been addressed by applying a random feature (RF) approximation \cite{rahimi2008random} to OMKL. In the the resulting method, named RF-based OMKL, the dimension of the optimization variables can be determined irrespective of the number of data $T$. Another advantage of RF-based OMKL is that learning a function can be solved using the powerful toolboxes from online convex optimization and online learning developed under vector spaces \cite{shen2019random}. Furthermore, an extension to active learning framework has been investigated in \cite{hong2020active}. Despite the success of kernel-based learning, the above methods cannot be applicable to distributed learning tasks caused in various IoT systems, which is the primary motivation of this work.

%
%

\vspace{0.1cm}
{\bf Contributions:} We consider the distributed learning setting, where measurements are made locally at each learner (e.g., IoT device) and the communication only occurs among neighboring learners. For this system, inspired by the success of kernel-based learning, we propose a distributed online learning framework with multiple kernels. The proposed method is named DOMKL. In the proposed DOMKL, each learner estimates its own function collaboratively with neighboring learners subject to the so-called consensus constraints. That is, neighboring learners aim at estimating an identical function by sharing their local information, while due to preserving a privacy, it is prohibitive to exchange local data directly. We contribute to this subject in the following ways:

\begin{itemize}
    \item  We extend the principles of the centralized OMKL \cite{shen2019random,hong2020active} into a distributed online learning framework. Thus, the proposed DOMKL also has the {\em scalability} in terms of the number of data, differently from the other multiple kernel learning methods, which makes it suitable for streaming data.
 
    \item In DOMKL, each kernel function at every learner is learned via online alternating direction method of multipliers (OADMM), by only exchanging the latest estimate with neighboring learners. Also, the weight for combining kernel functions is updated using the principle of hedge algorithm in a distributed way. We remark that by sharing gradients only, {\em privacy} at every learner is certainly ensured. 
    
    \item We theoretically prove that DOMKL can guarantee a sublinear regret $\Oc(\sqrt{T})$ with respect to both learning accuracy and constraint violation (i.e., discrepancy). We notice that the learning accuracy of DOMKL is in the same order at that of the centralized OMKL in \cite{shen2019random,hong2020active}. Specifically, our analysis reveals that a learned function at every learner almost achieves the performance of the best kernel function in the kernel dictionary, and has a negligible difference from those of its neighboring learners. 
    
    \item Via numerical tests with real datasets, we verify the effectiveness of the proposed DOMKL on regression and time-series prediction tasks, by showing that it achieves the almost same performance of the centralized OMKL.
\end{itemize}


%
%

The remaining part of this paper is organized as follows. In Section~\ref{sec:pre}, we briefly review the RF-based MKL, which is the baseline framework of our function learning.
In Section~\ref{sec:methods}, we describe the proposed distributed online learning framework with multiple kernels, named DOKL and DOMKL. Theoretical analysis is provided in Section~\ref{sec:TA} to verify the asymptotic optimality of the proposed methods. In Section~\ref{sec:Exp}, beyond the asymptotic analysis, we demonstrate the effectiveness of the proposed DOMKL via numerical tests with real datasets. Some concluding remarks are provided in Section~\ref{sec:con}.

{\em Notations:} Bold lowercase letters represent the column vectors. For any vector $\xv$, $\xv^{\trasp}$ denotes the transpose of $\xv$ and $\|\xv\|$ denote the $\ell_2$-norm of $\xv$. Also, $\langle\cdot,\cdot\rangle$ represents the inner product in Euclidean space. $\EE[\cdot]$ represents the expectation over an associated probability distribution. To simplify the notations, we let $[N]\eqdef\{1,...,N\}$ for any positive integer $N$. To clarify the notations, the $i$, $t$, $p$ indicate the indices of a learner, time, and kernel, respectively. 

%
%
\section{Preliminaries}\label{sec:pre}

We briefly review online multiple kernel learning based on random feature (RF) approximation (termed RF-based MKL), which will be used as the foundation of the proposed distributed online learning.  Given the training data $\{(\xv_1,y_1),...,(\xv_T,y_T)\}$, where $\xv_{t}\in \Xc \subseteq \RR^d$ and $y_t \in \Yc\subseteq \RR$, the objective of MKL is to learn a (non-linear) function $f: \Xc \rightarrow \Yc$ that minimizes the accumulate loss 
\begin{equation}
    \frac{1}{T}\sum_{t=1}^{T}\Lc(f(\xv_t),y_t),
\end{equation} where $f(\xv_t)$ and  $\Lc(\cdot,\cdot)$ represent an estimated label and a loss function, respectively. In a kernel-based learning \cite{gonen2011multiple, bazerque2013nonparametric, smola1998learning}, it is assumed that a target function $f(\xv)$ belongs to a reproducing Hilbert kernel space (RKHS) (denoted by $\Hc$), namely, $f(\xv)$ can be represented as
\begin{equation}
    f(\xv)=\sum_{t=1}^{\infty} \alpha_t \kappa(\xv,\xv_t),
\end{equation} 
where $\kappa(\xv,\xv_t): \Xc \times \Xc \rightarrow \Yc$ is a symmetric positive semidefinite basis function (called kernel), which measures the similarity between $\xv$ and $\xv_t$. Among various kernels, one representative example is the Gaussian kernel with a parameter $\sigma^2$, given as
\begin{equation} \label{eq:Gaussian_kernel}
    \kappa(\xv,\xv_t)= \exp\left(\frac{-\|\xv-\xv_t\|^2}{2\sigma^2}\right).
\end{equation} Also, a kernel is said to be {\em reproducing} if 
\begin{equation}
    \langle\kappa(\xv,\xv_{t_1}),\kappa(\xv,\xv_{t_2})\rangle_{\Hc} = \kappa(\xv_{t_1},\xv_{t_2}),
\end{equation} 
where $\langle\cdot,\cdot\rangle_{\Hc}$ denotes an inner product defined in the Hilbert space $\Hc$. The associated RKHS norm is defined as 
\begin{equation}\label{eq:kernel_fn}
    \|f\|_{\Hc}^2 \eqdef \sum_{t_1}\sum_{t_2}\alpha_{t_1}\alpha_{t_2}\kappa(\xv_{t_1},\xv_{t_2}).
\end{equation} 
The function learning problem over RKHS can be formulated:
\begin{equation}\label{eq:opt}
    \min_{f\in\Hc}\;\frac{1}{T} \sum_{t=1}^{T}\Lc(f(\xv_t),y_t).
\end{equation}
We remark that a loss function can be chosen in a task-specific way, e.g., least-square cost for regression and logistic cost for classification. Especially when the number of data is finite (e.g., $T$ training data), the representer theorem in \cite{wahba1990spline} shows that the optimal solution of (\ref{eq:opt}) is represented as
\begin{equation}\label{eq:appr}
    \hat{f}(\xv) = \sum_{t=1}^{T}\alpha_t \kappa(\xv,\xv_t).
\end{equation} The major drawback of this approach is the curse of dimensionality as the number of parameters $\alpha_t$'s (to be optimized) grows with the number of data $T$.

In \cite{rahimi2008random}, it has been addressed by introducing RF approximation for kernels. As in \cite{rahimi2008random}, the kernel $\kappa$ is assumed to be shift-invariant, i.e.,  $\kappa(\xv_{t_1},\xv_{t_2})=\kappa(\xv_{t_1} - \xv_{t_2})$ for any $t_1,t_2\in[T]$. Note that Gaussian, Laplacian, and Cauchy kernels satisfy the shift-invariance \cite{rahimi2008random}. For $\kappa(\xv_{t_1} - \xv_{t_2})$ absolutely integrable, its Fourier transform $\pi_{k}(\vv)$ exists and represents the power spectral density. Also, when $\kappa(\zerov)=1$ it can also be viewed as a probability density function (PDF). For a Gaussian kernel in (\ref{eq:Gaussian_kernel}), we have $\pi_{\kappa}(\vv)=\Nc(0,\sigma^{-2}\Id)$. Then, the kernel function can be rewritten as 
\begin{equation}
    \kappa(\xv_{t_1}-\xv_{t_2}) = \EE\left[\exp\left(j\vv^{\trasp}\left(\xv_{t_1}-\xv_{t_2}\right)\right)\right].
\end{equation}
Having a sufficient number of independent and identically distributed (i.i.d.) samples $\vv_i,\; i\in[D]$ from $\pi_{\kappa}(\vv)$, $\kappa(\xv_{t_1}-\xv_{t_2})$ can be well-approximated by the sample mean such as
\begin{equation}\label{RF_AP}
    \kappa(\xv_{t_1} - \xv_{t_2}) \approx \frac{1}{D}\sum_{i=1}^{D} \mbox{Re}\left(\exp\left(j\vv_{i}^{\trasp}\left(\xv_{t_1}-\xv_{t_2}\right)\right)\right),
\end{equation}where $\mbox{Re}(a)$ denotes the real part of a complex value $a$. Clearly, the accuracy of this approximation grows as the number of samples $D$ increases. In numerical tests, a proper $D$ will be chosen by considering the accuracy-complexity tradeoff. The approximation in (\ref{RF_AP}) can be rewritten as a vector form $\kappa(\xv_{t_1} - \xv_{t_2}) = \zv^{\trasp}(\xv_{t_1})\zv(\xv_{t_2})$,
where
\begin{equation}\label{eq:zv}
    \zv(\xv)=\frac{1}{\sqrt{D}}\left[\sin{\vv_1^{\trasp}\xv},...,\sin{\vv_{D}^{\trasp}(\xv),\cos{\vv_{1}^{\trasp}\xv},...,\cos{\vv_{D}^{\trasp}\xv}}\right]^{\trasp}.
\end{equation} Based on this, the optimal solution $\hat{f}(\xv)$ in (\ref{eq:appr}) can be well-approximated as
\begin{equation}\label{eq:RFA}
    \hat{f}(\xv) = \sum_{t=1}^{T}\alpha_t\zv^{\trasp}(\xv_t)\zv(\xv)  \eqdef \hat{\thetav}^{\trasp}\zv(\xv),
\end{equation} where the optimization variable $\hat{\thetav}$ is a $2D$-vector. Note that its dimension $2D$ can be determined irrespective of the number of data $T$.

RF-based kernel learning can be naturally extended into MKL framework, where a target function is formed as a linear (or convex) combination of multiple preselected kernels $\{\kappa_p: p\in[P]\}$. From \cite{micchelli2005learning}, the estimated function can be represented as
\begin{equation}\label{eq:MKL_f}
    \hat{f}(\xv)=\sum_{p=1}^{P} \hat{q}_p \hat{f}_p (\xv) \in \bar{\Hc},
\end{equation} where $\bar{\Hc}\eqdef \Hc_1 \bigotimes \Hc_2 \bigotimes \cdots \bigotimes \Hc_P$ and
$\hat{f}_p(\xv) \in \Hc_{p}$ which is a RKHS induced by the kernel $\kappa_p$, and $\hat{q}_p\in[0,1]$ denotes the combination weight of the associated kernel function $\hat{f}_p$. Also, under RF approximation, the kernel functions in (\ref{eq:MKL_f}) can be further simplified as 
\begin{equation}\label{eq:kernel_form}
    \hat{f}_{p}(\xv) = \hat{\thetav}_p^{\trasp} \zv_{p}(\xv),
\end{equation}
for $p\in[P]$, where $\zv_{p}(\xv)$ is defined in (\ref{eq:zv}) with $D$ number of i.i.d. samples from $\pi_{\kappa_p}(\vv)$. 
Assuming RF-based MKL, the function in (\ref{eq:kernel_form}), which is defined by the parameter $\hat{\thetav}_{p}$, will be assumed as the form of a learned (or estimated) function in the subsequent sections.

\begin{figure}[!t]
\centerline{\includegraphics[width=9cm]{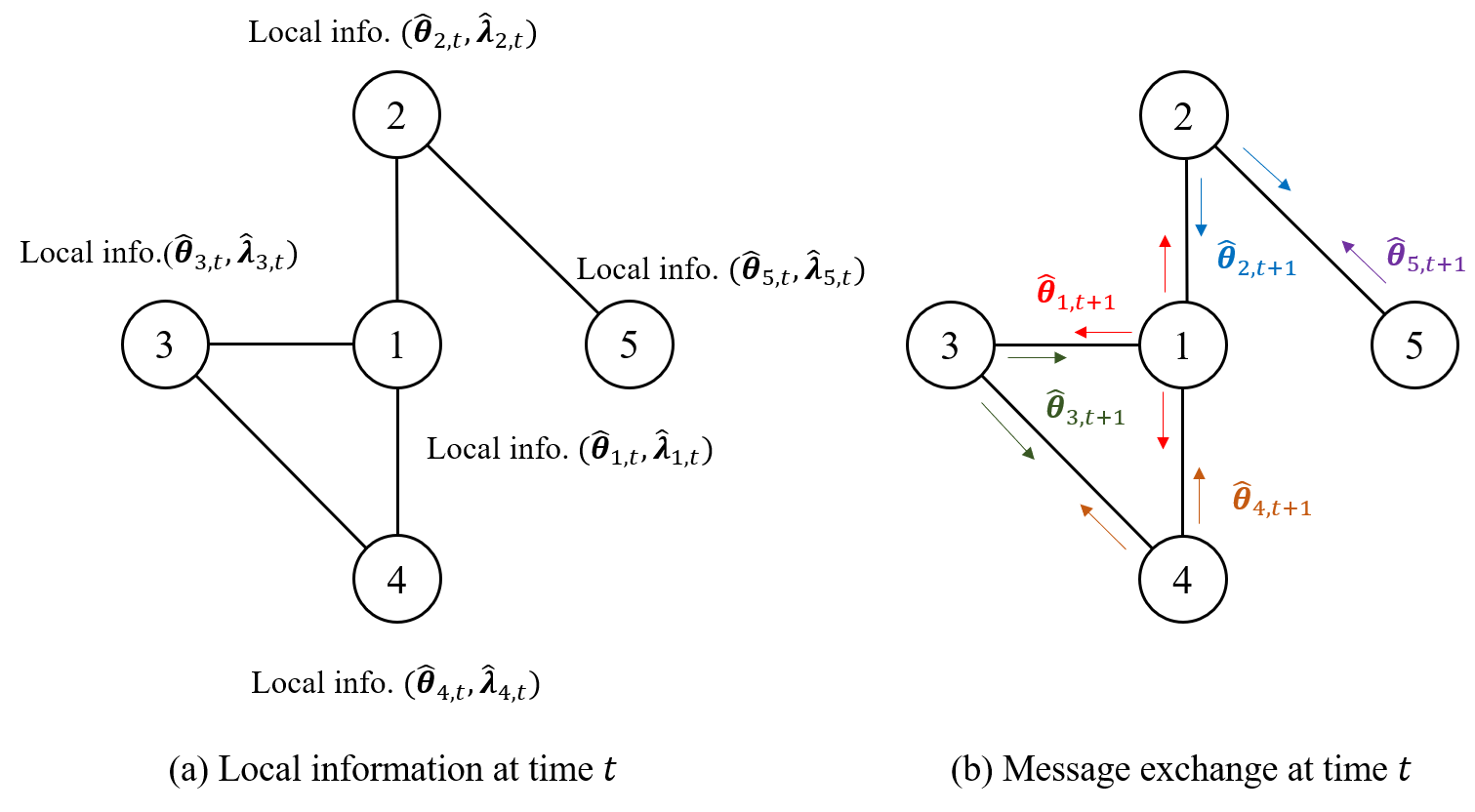}}
\caption{Description of the distributed online kernel learning (DOKL).}
\label{fig:OADMM}
\end{figure}

\section{Methods}\label{sec:methods}

Consider a distributed online learning task in a network with the set of learners (e.g., IoT devices) $\Vc=\{1,2,...,J\}$ interacting with each other over communication links. In particular, each learner aims at estimating a function in a sequential fashion, by leveraging its local streaming data and some information provided by neighboring learners. The connectivity of the $J$ learners is described by the set of unordered pairs of the learners (called edges), given as
\begin{equation}
    \Ec=\{\{i,j\}: \mbox{the learners } i \mbox{ and }  j \mbox{ are connected}\}.
\end{equation} Specifically, the learner $i$ can communicate with the learner $j$ only if $\{i,j\} \in \Ec$. Given the learner $j$, the index set of its neighboring learners is denoted as 
\begin{equation}
    \Nc_{j} = \{i: \{i,j\} \in \Ec\}\subseteq \Vc.
\end{equation} Also, our {\em communication model} is assumed as follows: Due to the privacy concerns, it is prohibitive to share local data with neighboring learners. Whereas, each learner $j$ can only transmit its latest estimated function (e.g., $\hat{\thetav}_{j,t+1}$) at time $t$ to the neighboring learners  $i \in \Nc_j$. Clearly, private local data should not be recovered from the estimated function $\hat{\thetav}_{j,t+1}$. For the example of Fig.~\ref{fig:OADMM}, we have that $\Vc=\{1,2,3,4\}$ and $\Ec=\{\{1,2\},\{1,3\},\{1,4\}, \{2,5\}, \{3,4\}\}$. Accordingly, we can obtain the set of neighbors as $\Nc_{1}=\{2,3,4\}$, $\Nc_{2}=\{1,5\}$, $\Nc_{3}=\{1,4\}$, $\Nc_{4}=\{1,3\}$, and $\Nc_{5}=\{2\}$. In this example, at time $t$, the learner $2$ transmits the estimated parameter $\hat{\thetav}_{2,t+1}$ to the learners 1 and 5, and receives the $\hat{\thetav}_{1,t+1}$ and $\hat{\thetav}_{5,t+1}$ from them, respectively.

Focusing on the above model, we propose a distributed online learning algorithm with the RF-based kernel in Section~\ref{subset:single}, and extend it into a multiple kernel framework in Section~\ref{subset:multiple}.


%
%
\subsection{The Proposed DOKL (Single Kernel)}\label{subset:single}

We first define the problem setting of a distributed kernel learning (DKL) and then extend it into an online framework. In the DKL, each learner $j$ has its own local data $\{(\xv_{j,1},y_{j,1}),...,(\xv_{j,T},y_{j,T})\}$. Leveraging them and some information from neighboring learners, the learner $j$ estimates a  (non-linear) function $\hat{f}_{j}(\xv)$. Following the RF approximation in (\ref{eq:RFA}), it is fully characterized by a $2D$-vector $\hat{\thetav}_{j}$  as
\begin{equation}\label{eq:single_k}
    \hat{f}_{j}(\xv) = \hat{\thetav}_{j}^{\trasp}\zv(\xv), 
\end{equation} where $\zv(\xv)$ is defined in (\ref{eq:zv}). Given the connectivity (or communication model) $\Ec$, the objective of DKL is to solve the following optimization problem in a distributed way:
\begin{align}
    &\argmin_{\thetav} \sum_{t=1}^T \sum_{j=1}^{|\Vc|} \Lc\left(\thetav_j^{\trasp}\zv(\xv_{j,t}), y_{j,t}\right)\nonumber\\
    &\mbox{ subject to } \thetav_i = \thetav_j, \;\; \forall \{i,j\}\in \Ec,\label{eq:const_o1}
\end{align}  where $\thetav=[\thetav_1;\thetav_2;\cdots;\thetav_{|\Vc|}]$ with $\thetav_j \in \RR^{2D\times 1}$. In \cite{boyd2011distributed}, it was shown that the above problem can be efficiently solved via ADMM. Toward this, the optimization problem in (\ref{eq:const_o1}) can be reformulated as
\begin{align}
    &\argmin_{\thetav} \sum_{t=1}^T \sum_{j=1}^{|\Vc|} \Lc\left(\thetav_j^{\trasp}\zv(\xv_{j,t}), y_{j,t}\right)\\
    &\mbox{ subject to } \Am\thetav + \Bm\gammav = \zerov,\label{eq:const_o2}
\end{align} for some matrices $\Am \in \RR^{4D|\Ec|\times 2D|\Vc|}$ and $\Bm\in\RR^{4D|\Ec|\times 2D|\Ec|}$, where each component of an auxiliary vector $\gammav$ is assigned to an edge such as
\begin{equation}
    \gammav=\left[\gammav_{\{i,j\}}: \{i,j\} \in \Ec\right] \in \RR^{2D|\Ec|\times 1}.
\end{equation} Using the connectivity $\Ec$, the matrices $\Am$ and $\Bm$ can be obtained straightforwardly. Consider the example of $\Vc=\{1,2,3\}$ and $\Ec=\{\{1,2\},\{1,3\}\}$. In this case, the matrices of $\Am$ an $\Bm$ are determined as
\begin{align}
    \Am &=\left[
    \begin{matrix} \Id & \zerov & \zerov \\
    \Id & \zerov & \zerov\\
    \zerov &\Id & \zerov\\
    \zerov&\zerov&\Id
    \end{matrix}\right] \mbox{ and }\Bm =\left[\begin{matrix}
    -\Id & \zerov\\
    \zerov & -\Id\\
    -\Id & \zerov\\
    \zerov & -\Id
    \end{matrix}
    \right],
\end{align} where $\Id$ and $\zerov$ represent the $2D\times 2D$ identify and zero matrices, respectively. It is easily verified that the constraint in (\ref{eq:const_o2}) is equivalent to that in (\ref{eq:const_o1}), i.e.
\begin{equation}\label{eq:example_A}
    \Am\thetav + \Bm\gammav = \left[\begin{matrix} \thetav_1 - \gammav_{\{1,2\}}\\
    \thetav_1 - \gammav_{\{1,3\}}\\
    \thetav_2 - \gammav_{\{1,2\}}\\
    \thetav_3 - \gammav_{\{1,3\}}
    \end{matrix}
    \right] = \zerov.
\end{equation}



We are now ready to explain the proposed {\em distributed online kernel learning} (DOKL). The primary goal of DOKL is to solve the optimization problem (\ref{eq:const_o2}) in an  online fashion. For the centralized setting \cite{hong2020active}, the online kernel learning problem has been efficiently solved via online gradient descent (OGD). Furthermore, it was shown that this approach can achieve the optimal sublinear regret $\Oc(\sqrt{T})$ \cite{hong2020active}. For the distributed setting, we will solve it using online ADMM (OADMM) proposed in \cite{wang2013online} and show that the optimal sublinear regret is also achieved.

At every time $t$, each learner $j$ observes the incoming data $(\xv_{j,t},y_{j,t})$ and optimizes the local function $\hat{f}_{j,t+1}$ (defined by a parameter $\hat{\thetav}_{j,t+1}$) via OADMM. Specifically, at every time $t$, OADMM solves the following regularized optimization problem:
\begin{align}
    \hat{\thetav}_{t+1} &= \argmin_{\thetav}  \sum_{j=1}^{|\Vc|} \Lc\left(\thetav_j^{\trasp}\zv(\xv_{j,t}), y_{j,t}\right) + \frac{\eta}{2}\left\|\thetav - \hat{\thetav}_t\right\|^2\nonumber\\
    &\mbox{subject to } \Am\thetav + \Bm \gammav = 0, \label{eq:opt1}
\end{align} where $\thetav=[\thetav_1;\thetav_2;\cdots;\thetav_{|\Vc|}]$ with $\thetav_j \in \RR^{2D\times 1}$. Accordingly, the augmented Lagrangian of (\ref{eq:opt1}) at time $t$  is given as
\begin{align}
    L_{\rho}^t(\thetav,\gammav, \lambdav) &\eqdef \sum_{j=1}^{|\Vc|} \Lc\left(\thetav_j^{\trasp}\zv(\xv_{j,t}), y_{j,t}\right) + \lambdav^{\transp}\left(\Am\thetav+\Bm\gammav\right) \nonumber\\
    &\;\;\; + \frac{\eta}{2}\left\|\thetav-\hat{\thetav}_t\right\|^2 + \frac{\rho}{2}\left\|\Am\thetav+\Bm\gammav_t\right\|^2.
\end{align} Thus, at time $t$, OADMM has the following update steps:
\begin{align}
    \hat{\thetav}_{t+1}&=\argmin_{\thetav} \sum_{j=1}^{|\Vc|} \Lc\left(\thetav_j^{\trasp}\zv(\xv_{j,t}), y_{j,t}\right) + \hat{\lambdav}_t^{\trasp} \left(\Am\thetav+\Bm\hat{\gammav}_t\right) \nonumber\\
    &\;\;\;\;\;\;\;\;\;\;\;\; + \frac{\rho}{2}\left\|\Am\thetav+\Bm\hat{\gammav}_t\right\|^2 + \frac{\eta}{2}\left\|\thetav-\hat{\thetav}_t\right\|^2,\label{eq:theta_j}\\
    \hat{\gammav}_{t+1} 
    &=\argmin_{\gammav}\;\; \hat{\lambdav}_t^{\transp}\left(\Am\hat{\thetav}_{t+1}+\Bm\gammav\right) +\frac{\rho}{2}\left\|\Am\hat{\thetav}_{t+1}+\Bm\gammav\right\|^2,\label{eq:zv}\\
    \hat{\lambdav}_{t+1}&=\hat{\lambdav}_t + \rho\left(\Am\hat{\thetav}_{t+1}+\Bm\hat{\gammav}_{t+1}\right).\label{eq:lambdav}
\end{align} 
Leveraging the structures of $\Am$ and $\Bm$, we will show that the above update steps can be performed in a distributed way, by only exchanging the latest estimates $\hat{\thetav}_{j,t}$'s among neighboring learners. The corresponding key lemma is given in the below.

%
%
\vspace{0.1cm}
\begin{lemma}\label{lem1-OADMM} The optimization problems in (\ref{eq:theta_j})-(\ref{eq:lambdav}) can be solved in a distributed way, where each learner $j$ only exchanges the estimate  $\hat{\thetav}_{j,t+1}$ with the neighboring learners $i\in\Nc_j$. The corresponding updates at the learner $j$ are derived as follows:
\begin{align}\hspace{-0.1cm}
    &\hat{\thetav}_{j,t+1}=\argmin_{\thetav_j} \Lc\left(\thetav_{j}^{\trasp}\zv(\xv_{j,t}), y_{j,t}\right) + \hat{\lambdav}_{j,t}^{\transp}\thetav_j \nonumber\\
    &\;\;\;\;\;\;\;+\frac{\rho}{2}\sum_{i\in\Nc_j}\Big\|\thetav_j - \frac{\hat{\thetav}_{j,t}+\hat{\thetav}_{i,t}}{2}\Big\|^2 + \frac{\eta}{2}\left\|\thetav_j - \hat{\thetav}_{j,t}\right\|^2,\label{eq:theta-update}\\
  &\hat{\lambdav}_{j,t+1} =\hat{\lambdav}_{j,t} + \frac{\rho}{2}\sum_{i\in\Nc_j}\left(\hat{\thetav}_{j,t+1}-\hat{\thetav}_{i,t+1}\right). \label{eq:lambda-update}
\end{align} 
\end{lemma}
\begin{IEEEproof} The proof is provided in Appendix~\ref{app:lem1}.
\end{IEEEproof}
\vspace{0.2cm}

From Lemma~\ref{lem1-OADMM}, every learner in the network operates in the following way:
\begin{itemize}
    \item At every time $t$, each learner $j$ in the network is aware of the latest estimates $\hat{\thetav}_{j,t}$ and $\hat{\lambdav}_{j,t}$ from the previous iteration. 
    \item Using them, it locally optimizes $\hat{\thetav}_{j,t+1}$ by solving the convex optimization in (\ref{eq:theta-update}). See Remark~\ref{rem:opt} for the case of quadratic loss function.
    \item Each learner $j$ transmits the $\hat{\thetav}_{j,t+1}$ to and receives $\{\hat{\thetav}_{i,t+1}: i \in \Nc_{j}\}$ from the neighboring learners.
    \item Using them, each learner $j$ updates the discrepancy $\hat{\lambdav}_{j,t+1}$ locally.
\end{itemize} Focusing on the learner $j$, the above procedures are described in {\bf Algorithm 1}.

\vspace{0.1cm}
\begin{remark}\label{rem:opt} Suppose that the {\em quadratic} loss function is used:
\begin{align}
    \Lc\left(\thetav^{\transp}\zv(\xv_t),y_t\right) = \left(\thetav^{\transp}\zv(\xv_{j,t})-y_{j,t}\right)^2.
\end{align} In this case, the learner $j$ can simply find the optimal solution of (\ref{eq:theta-update}).
Then, the learner $j$ can simply find the optimal solution of 
 (\ref{eq:theta-update})  as
\begin{align}
    \hat{\thetav}_{j,t+1} &= \Big(2\zv(\xv_{j,t})\zv(\xv_{j,t})^{\transp} + (\eta+\rho|\Nc_j|)\Id\Big)^{-1}\nonumber\\
    &\;\;\;\;\; \cdot\left(2y_{j,t}\zv(\xv_{j,t}) +\eta\hat{\thetav}_{j,t} + \rho\hat{\gammav}_{j,t} - \hat{\lambdav}_{j,t}\right),
\end{align} where $\hat{\gammav}_{j,t}=\sum_{i\in \Nc_j} (\hat{\thetav}_{j,t}+\hat{\thetav}_{i,t})/2$. The above closed-form expression will be used for experimental tests in Section~\ref{sec:Exp}.
\end{remark}

\begin{algorithm}
\caption{DOKL (at the leaner $j$)}
\begin{algorithmic}[1]

\State {\bf Input:} Network $(\Vc,\Ec)$, a kernel $\kappa$, parameters ($\rho$, $\eta$),  the number of random features $D$ (for RF approximation).  
\State {\bf Output:} A sequence of functions $\hat{f}_{j,t}(\xv),\; t\in [T+1]$.
\vspace{0.05cm}
\State {\bf Initialization:} $\hat{\thetav}_{j,1} = \zerov$ and $\hat{\lambdav}_{j,1} = \zerov$.
\vspace{0.05cm}
\State {\bf Iteration:} $t=1,...,T$
\begin{itemize}
    \item Receive a streaming data $(\xv_{j,t},y_{j,t})$.
    \item Construct $\zv(\xv_{j,t})$ via (\ref{eq:zv}) using the kernel $\kappa$.
    \item Update $\hat{\thetav}_{j,t+1}$ from (\ref{eq:theta-update}). 
    \item Transmit $\hat{\thetav}_{j,t+1}$ to and receive $\hat{\thetav}_{i,t+1}$ from the neighboring learners $i \in \Nc_j$.
    \item Update $\hat{\lambdav}_{j,t+1}$ from (\ref{eq:lambda-update}).
\end{itemize}
\end{algorithmic}
\end{algorithm}


%
%
\subsection{The Proposed DOMKL (Multiple Kernels)}\label{subset:multiple}

As an extension of the previous section, we will present a distributed online learning with {\em multiple kernels} (termed DOMKL). This approach can address the challenging problem of choosing a proper kernel in DOKL. Throughout the paper, it is assumed that there are $P$ kernels in a kernel dictionary, each of which is defined by $\kappa_{p}$ for $p\in[P]$. In the proposed DOMKL, every learner follows the same procedures separately. Focusing on the operations of the learner $j$, thus,  DOMKL consists of the following two steps:

\vspace{0.1cm}
{\bf i) Local step:} This step learns a set of single kernel functions $\hat{f}_{j,t+1}^{p}(\xv) \in \Hc_{p}$ for $p \in[P]$. The learner $j$ observes the incoming data $(\xv_{j,t}, y_{j,t})$ and optimizes the local functions $\hat{f}_{j,t+1}^{p}(\xv)$ for $p \in [P]$, via online optimization.   Following the RF approximation in (\ref{eq:RFA}), each kernel function is determined by the $2D$-vector $\hat{\thetav}_{j,t+1}^{p}$ such as
\begin{equation}
    \hat{f}_{j,t+1}^{p}(\xv) = \left(\hat{\thetav}_{j,t+1}^{p}\right)^{\trasp}\zv_{p}(\xv), 
\end{equation} where $\zv_{p}(\xv)$ is defined in (\ref{eq:zv}). As in Section~\ref{subset:single}, the parameter vector  $\hat{\thetav}_{j,t+1}^{p}$ is optimized in an online fashion via OADMM. The corresponding update rules are provided in (\ref{eq:theta-update}) and (\ref{eq:lambda-update}). We remark that each kernel function $\hat{f}_{j,t+1}^{p}$ is optimized independently from the other kernel functions. Namely, ADMM is performed for each kernel separately.  During this step, the learner $j$ obtained the $P$ kernel functions defined by the parameters
$\{\hat{\thetav}_{j,t+1}^{p}: p \in [P]\}$.

\vspace{0.1cm}
{\bf ii) Global step:} In this step, every learner $j$ seeks the best function approximation $\hat{f}_{j,t+1}(\xv)$ by combining its local kernel functions $\{\hat{f}_{j,t+1}^{p}(\xv), p\in [P]\}$ with proper weights $\{\hat{q}_{j,t+1}^{p}, p\in[P]\}$:
\begin{equation}\label{eq:DOMKL_f}
    \hat{f}_{j,t+1}(\xv) = \sum_{p=1}^{P} \hat{q}_{j,t+1}^{p}\hat{f}_{j,t+1}^{p}(\xv),
\end{equation} where $\sum_{p=1}^P \hat{q}_{j,t+1}^{p} = 1$ with $\hat{q}_{j,t+1}^{p} \in [0,1]$. Here, the weights can capture the reliabilities (or accuracy) of the corresponding kernel functions. Following the online learning framework \cite{bubeck2011introduction}, the weights are determined by the so-called {\em exponential strategy} (EXP strategy) in which they are determined on the basis of the past losses (i.e., the reliabilities of $P$ kernels) as follows. 
We first define the {\em local} accumulated losses:
\begin{equation}
    \hat{w}_{j,t+1}^{p} = \exp\left(-\frac{1}{\eta_g} \sum_{\tau=1}^t \Lc\left(\hat{f}_{j,\tau}^{p}(\xv_{j,\tau}), y_{j,\tau}\right) \right),\label{eq:local-w}
\end{equation}for $p\in P$. Then, using $\{\hat{w}_{i,t+1}^{p}: i \in \Nc_j\}$ (obtained from the neighboring learners), the weights are defined as
\begin{equation}
    \hat{q}_{j,t+1}^{p} = \frac{\hat{w}_{j,t+1}^{p}\cdot \prod_{i\in\Nc_j}\hat{w}_{i,t+1}^{p}}{\sum_{p=1}^P \Big(\hat{w}_{j,t+1}^{p}\cdot \prod_{i\in\Nc_j}\hat{w}_{i,t+1}^{p}\Big)},\label{eq:weight-update}
\end{equation} for some parameter $\eta_g>0$. Notice that from the above update rule, the weights of the learner $j$ are determined by the local data of the neighboring learners as well as its own local data, i.e.,
\begin{align*}
    \hat{w}_{j,t+1}^{p}\cdot \prod_{i\in\Nc_j}\hat{w}_{i,t+1}^{p} = \exp\left(-\frac{1}{\eta_g}L_{j,t+1}\right),
\end{align*} where the accumulated losses are obtained as
\begin{align*}
    L_{j,t+1} &= \sum_{\tau=1}^{t} \Lc\left(\hat{f}_{j,\tau}^{p}(\xv_{j,\tau}), y_{j,\tau}\right) \\
    &\;\;\;\;\;\;\;\;\;\;\;\;\;\;\;\;\;\;\;\;\;\;\;\;+\sum_{i\in \Nc_j} \sum_{\tau=1}^{t} \Lc\left(\hat{f}_{i,\tau}^{p}(\xv_{i,\tau}), y_{i,\tau}\right).
\end{align*} Here, the second term can capture the reliability of the kernel function $p$ obtained from the data of neighboring learners.

\vspace{0.1cm}
\begin{remark}
In particular when a network forms an {\em acyclic graph}, the weights can be further elaborated using the principle of message-passing \cite{mezard2009information}. In this case, the weights (i.e., the reliabilities of $P$ kernels) can be updated using the local data of connected learners, in addition to that of neighboring learners. Here, we say that two learners are connected with a length $\zeta$ if there exists a path of length $\zeta$ between the two learners. Also, we let $\Nc_{j}^{(\zeta)}$ denote the index set containing all the length-$\zeta$ connected learners from the learner $j$, i.e., $\Nc_{j}^{(\zeta)}=\{i\in\Vc: \mbox{there exists a length-$\zeta$ path} \mbox{between the learners $j$ and $i$}\}$
with $\Nc_{j}^{(1)} = \Nc_j$. Based on the principle of message-passing, the message transmitted from the learner $j$ to the neighboring learner $i \in \Nc_j$ is determined as
\begin{equation}
    m_{j\rightarrow i,t+1} = \hat{w}_{j,t+1}^{p}\cdot \prod_{\ell \in \Nc_j: \ell\neq i} m_{\ell\rightarrow j, t}
\end{equation} with initial values $m_{\ell \rightarrow j,1}=1$ for all $\{\ell,j\}\in \Ec$. Accordingly, the weights of the learner $j$ are updated from the incoming messages $\{m_{\ell\rightarrow j, t}: \ell \in \Nc_{j}\}$ such as
\begin{equation}
    \hat{q}_{j,t+1}^{p} = \frac{\hat{w}_{j,t+1}^{p}\cdot\prod_{\ell \in \Nc_j} m_{\ell\rightarrow j, t}}{\sum_{p=1}^P \left(\hat{w}_{j,t+1}^{p}\cdot \prod_{\ell \in \Nc_j} m_{\ell\rightarrow j, t} \right) },
\end{equation} for some parameter $\eta_g>0$. This update rule can ensure that local information for weights (i.e., the reliabilities of the $P$ kernels) can be propagated over the network as a time (or iteration) $t$ grows, which is shown in the below:
\begin{equation}
    \hat{w}_{j,t+1}^{p}\cdot \prod_{\ell \in \Nc_j} m_{\ell\rightarrow j, t} = \exp\left(-\frac{1}{\eta_g}L_{j, t+1}\right),
\end{equation} where the accumulated losses are given as
\begin{align}
     L_{j, t+1} &= \sum_{\tau=1}^{t} \Lc\left(\hat{f}_{j,\tau}^{p}(\xv_{j,\tau}), y_{j,\tau}\right)\nonumber\\
     &\;\;\;\;\;\;\;\;\;\;\; +\sum_{\zeta=1}^{t-1}\sum_{i\in \Nc_j^{(\zeta)}} \sum_{\tau=1}^{t-\zeta} \Lc\left(\hat{f}_{i,\tau}^{p}(\xv_{i,\tau}), y_{i,\tau}\right).
\end{align} The above update philosophy can be maintained even for cyclic graphs, as long as the girth of a network graph is larger than the overall number of iterations $T$. Otherwise, some losses can be reflected due to the duplication.
\end{remark}


Note that the learned function in (\ref{eq:DOMKL_f}) will be used to estimate the label $\hat{y}_{j,t+1}$ of an incoming local data $\hat{\xv}_{j,t+1}$. The detailed procedures of DOMKL are provided in {\bf Algorithm 2}. 


\begin{algorithm}
\caption{DOMKL (at the leaner $j$)}
\begin{algorithmic}[1]
\State {\bf Input:} Kernels $\kappa_p$, $p\in[P]$, parameters $(\eta, \eta_g)$,  the number of random features $D$ (for RF approximation).  
\State {\bf Output:} A sequence of functions $\hat{f}_{j,t}(\xv)$, $t\in [T+1]$.
\State {\bf Initialization:}  $\hat{\thetav}_{j,1}^{p} = \zerov$, $\hat{\lambdav}_{j,1}^p=\zerov$ and $\hat{w}_{j,1}^p = 1, \forall p\in[P]$.
\State {\bf Iteration:} $t=1,...,T$

\hspace{-0.8cm}$\bullet$ Receive a streaming data $(\xv_{j,t},y_{j,t})$.

\hspace{-0.8cm}$\bullet$ Construct $\zv_{p}(\xv_{j,t})$ via (\ref{eq:zv}) using the kernel $\kappa_p, p\in[P]$.

\hspace{-0.8cm}$\bullet$ {\bf Local step}: 
\begin{itemize}
   \item[$-$]  Update $\hat{\thetav}_{j,t+1}^{p}$ from (\ref{eq:theta-update}) for $p\in[P]$.
   \item[$-$] Transmit $\{\hat{\thetav}_{j,t+1}^{p}: p\in[P]\}$ to and receive $\{\hat{\thetav}_{i,t+1}^{p}: p\in[P]\}$ from the neighboring learners $i \in \Nc_j$.
   \item[$-$] Update $\hat{\lambdav}_{j,t+1}^{p}$ from (\ref{eq:lambda-update}).
\end{itemize}

\hspace{-0.8cm}$\bullet$ {\bf Global step}:
\begin{itemize}
    \item[$-$] Update $\hat{w}_{j,t+1}^{p}$ via (\ref{eq:local-w}).
    \item[$-$] Transmit $\{\hat{w}_{j,t+1}^{p}: p\in [P]\}$ and receive $\{\hat{w}_{i,t+1}^{p}: p\in [P]\}$ with the neighboring learners $i \in \Nc_j$.
    \item[$-$] Update $\hat{q}_{j,t+1}^{p}$ via (\ref{eq:weight-update}).
    \item[$-$] Update $\hat{f}_{j,t+1}(\xv) = \sum_{p=1}^{P} \hat{q}_{j,t+1}^{p}\left(\hat{\thetav}_{j,t+1}^{p}\right)^{\transp}\zv_{p}(\xv)$.
    \end{itemize}
\end{algorithmic}
\end{algorithm}

%
%
\section{Theoretical Analysis}\label{sec:TA}

We analyze the performances of the proposed DOKL and DOMKL in terms of the cumulative regrets for both learning accuracy and constraint violation (i.e., discrepancy). We let $\hat{f}_{j,t}$ denote the estimated function of the learner $j$. Let $f_{j}^{\star}$ denote the optimal function for the data of the learner $j$, i.e.,
\begin{equation}
    f_{j}^{\star} = \argmin_{f} \sum_{t=1}^{T}\Lc\left(f(\xv_{j,t}),y_{j,t}\right).
\end{equation} Then, the cumulative regrets for learning accuracy and discrepancy at the leaner $j$ are formally defined as
\begin{align*}
    {\rm regret}_{\rm a}^j(T) &=\sum_{t=1}^{T} \Lc\left(\hat{f}_{j}^t(\xv_{j,t}),y_{j,t}\right)-\Lc(f_{j}^{\star}\left(\xv_{j,t}),y_{j,t}\right)\\
    {\rm regret}_{\rm d}^j (T) &=\sum_{t=1}^{T} \left|\sum_{i\in \Nc_j}\left(\hat{f}_{j,t}(\xv_{j,t}) -\hat{f}_{i,t}(\xv_{j,t})\right) \right|^2.
\end{align*}

From now on, we will prove that the proposed DOMKL and DOMKL can achieve the sublinear regrets $\Oc(\sqrt{T})$ for both learning accuracy and discrepancy. When $T\rightarrow \infty$, thus, the regret per time will be disappeared. Recall that under the RF approximation, each kernel function $\hat{f}_{j,t}^{p}$ has the form of 
\begin{equation}
    \hat{f}_{j,t}^{p}(\xv) = \left(\hat{\thetav}_{j,t}^{p}\right)^{\trasp}\zv_{p}(\xv).
\end{equation} For ease of exposition, given the data $(\xv_{j,t}, y_{j,t})$, the loss function with respect to $\thetav$ is represented as 
\begin{equation}
    \Lc\left(\thetav^{\trasp}\zv_p(\xv_{j,t}),y_{j,t}\right) \eqdef \Lc_{j,t}^{p}(\thetav).
\end{equation} Throughout the paper, the above two notations will be used interchangeably. Also, we let $\thetav_{j,p}^{\star}$ denote the optimal RF approximation function at the kernel $p$, i.e.,
\begin{equation}
    \thetav_{j}^{\star} = \argmin_{\thetav} \sum_{t=1}^{T}\Lc\left(\thetav^{\trasp}\zv_{p}(\xv_{j,t}),y_{j,t}\right), \;\forall j\in\Vc,
\end{equation} with the consensus constraint in (\ref{eq:opt1}).
For the analysis, the following conditions are assumed:
\begin{itemize}
    \item {\bf (a1)} For any fixed $\zv_{p}(\xv_{j,t})$ and $y_{j,t}$, the loss function $\Lc_{j,t}^{p}(\thetav)$ is convex with respect to $\thetav$, differentiable, and bounded as  $\Lc_{j,t}^p(\thetav) \in [0,L_u]$. Also, its gradient is bounded, i.e., $\|\nabla\Lc_{j,t}^{p}(\thetav)\|^2 \leq G$ for some positive constant $G$.
    \item {\bf (a2)} For any kernel $\kappa_p$, $\hat{\thetav}_{j,t}^{p}$ belongs to a bounded set $\Theta_p \subseteq \RR^{2D\times 1}$, i.e., $\|\hat{\thetav}_{j,t}^p\|^2\leq C$  for some positive constant $C$.
    \item {\bf (a3)} For any $t \in T$, $\Lc_{j,t}^{p}(\hat{\thetav}_{j,t}^{p}) - \Lc_{j,t}^{p}(\thetav_{j}^{\star}) \geq  -  B$
    for some positive constant $B$.
\end{itemize} The assumptions {\bf (a1)} and {\bf (a2)} are in general required in the online learning setting \cite{shen2019random, hazan2016introduction, wang2013online, hong2020active}. Also, the assumption {\bf (a3)} is required to prove the sublinear regret of constraint violation (i.e., discrepancy), which is true if convex functions are bounded from below or Lipschitz continuous \cite{wang2013online}. 
Additionally, for the proof of Theorem~\ref{thm2}, the following assumption is required:
\begin{itemize}
    \item {\bf (a4)} For any fixed $j \in \Vc$ and $p \in [P]$, there exists a sequence of $\epsilon_t$'s such that $|\hat{q}_{j,t}^p - \hat{q}_{i,t}^p|\leq \epsilon_t$ and $\sum_{t=1}^T \epsilon_t^2 \leq \Oc(\sqrt{T})$ for $\forall i \in\Nc_j$, where $\hat{q}_{j,t}^p$ is defined in (\ref{eq:weight-update}).
\end{itemize} In fact, the upper-bound $\epsilon_t$ is determined on the basis of network structure (i.e., the connectivity of learners in the network). For example, when the network is a complete graph, we can easily obtain the $\epsilon_t = 0$ for all $t \in [T]$. Obviously, as the connectivity of a network becomes sparse, $\epsilon_t$ tends to increase. Via numerical tests, we have confirmed that the assumption {\bf (a4)} might not be tight, i.e., it can be easily satisfied in practical network structures.


We first state the main results of this section in Theorems~\ref{thm1} and~\ref{thm2} below, and the proofs will be provided in Sections~\ref{subsec:proof1} and~\ref{subsec:proof2}.
%
%
\vspace{0.1cm}
\begin{theorem}\label{thm1} Under the assumptions {\bf (a1)} - {\bf (a3)}, DOKL in Algorithm 1 with the parameters $\rho=\eta=\sqrt{T}$ and a kernel $\kappa_p$ can achieve the sublinear regrets  as
\begin{align*}
    &{\rm regret}_{\rm a}^{j}(T)=\sum_{t=1}^{T} \Lc\left( \hat{\thetav}_{j,t}^{\trasp}\zv_{p}(\xv_{j,t}),y_{j,t}\right)\\  
    &\;\;\;\;\;\;\;\;\;\;\;\;\;\;\;\;\;\;\;\;\;\;\;\;\;\;\;\;\;\;\;\;\;\;\;- \Lc\Big((\thetav_{j}^{\star})^{\trasp}\zv_{p}(\xv_{j,t}),y_{j,t}\Big) \leq \Oc(\sqrt{T})\\
    &{\rm regret}_{\rm d}^{j}(T)\\
    &\;\;\;\;\;\;\;\; =\sum_{t=1}^{T} \left|\sum_{i\in\Nc_j}\left(\hat{\thetav}_{j,t}^{\trasp}\zv_{p}(\xv_{j,t})-  \hat{\thetav}_{i,t}^{\trasp}\zv_{p}(\xv_{j,t})\right)\right|^2 \leq \Oc(\sqrt{T}).
\end{align*}
\end{theorem}

\begin{theorem}\label{thm2} Under the assumptions {\bf (a1)} - {\bf (a4)}, DOMKL in Algorithm 2 with the parameters $\rho=\eta=\eta_g=\sqrt{T}$ and kernels $\{\kappa_p: p\in[P]\}$ can achieve the sublinear regrets as
\begin{align*}
    &{\rm regret}_{\rm a}^{j}(T) = \sum_{t=1}^{T} \Lc\left(\sum_{p=1}^P \hat{q}_{j,t}^p\hat{\thetav}_{j,t}^p\zv_{p}(\xv_{j,t}), y_{j,t}\right) \\
    &\;\;\;\;\;\;\;\;\;\;\;\;\;\;\;\;\;\;\;\;\;\;- \min_{1\leq p\leq P}\Lc\left(\left(\thetav_{j}^{\star}\right)^{\trasp}\zv_{p}(\xv_{j,t}),y_{j,t}\right)\leq \Oc(\sqrt{T})\\
    &{\rm regret}_{\rm d}^{j}(T) =\sum_{t=1}^{T} \left|\sum_{i\in\Nc_j}\left(\sum_{p=1}^{P}\hat{q}_{j,t}^p\left(\hat{\thetav}_{j,t}^p\right)^{\trasp}\zv_{p}(\xv_{j,t})\right.\right.\\
    &\;\;\;\;\;\;\;\;\;\;\;\;\;\;\;\;\;\;\;\;\;\;\;\;\;\;\left.\left.-  \sum_{p=1}^{P}\hat{q}_{i,t}^p\left(\hat{\thetav}_{i,t}^p\right)^{\trasp}\zv_{p}(\xv_{j,t})\right)   \right|^2 \leq \Oc(\sqrt{T}).
\end{align*} 
\end{theorem}

\vspace{-0.1cm}
\subsection{Proof of Theorem 1}\label{subsec:proof1}

Let $\hat{\thetav}_{j,t}$ and $\hat{\lambdav}_{j,t}$ be the output of the proposed DOKL (in {\bf Algorithm 1}). Using them, we first derive the useful lemma:


\begin{lemma}\label{lem2} Letting $\hat{\gammav}_{j,t} = \sum_{i\in\Nc_j}(\hat{\thetav}_{j,t}+\hat{\thetav}_{i,t})/2$, we have the following upper-bound:
\begin{align}
    &\Lc_{j,t}^p(\hat{\thetav}_{j,t+1}) - \Lc_{j,t}^p \left(\thetav_{j}^{\star}\right)\nonumber \\
    &\leq  \frac{1}{2\rho} \left(\|\hat{\lambdav}_{j,t}\|^2-\|\hat{\lambdav}_{j,t+1}\|^2\right)-\frac{\rho}{2}\|\hat{\thetav}_{j,t+1}-\hat{\gammav}_{j,t}\|^2\nonumber\\
    &+\frac{\eta}{2}\left(\|\hat{\thetav}_{j,t}-\thetav_{j}^{\star}\|^{2} - \|\hat{\thetav}_{j,t+1}-\thetav_{j}^{\star}\|^{2}-\|\hat{\thetav}_{j,t+1}-\hat{\thetav}_{j,t}\|^2\right)\nonumber\\
    &+\frac{\rho}{2}\left(\|\hat{\gammav}_{j,t}-\thetav_{j}^{\star}\|^2 - \|\hat{\gammav}_{j,t+1}-\thetav_{j}^{\star}\|^2\right).\label{eq:proof0}
\end{align} 
\end{lemma}

\begin{IEEEproof} The proof is provided in Appendix~\ref{app:lem2}.
\end{IEEEproof}

We are now ready to prove the main results of  Theorem~\ref{thm1}.

\vspace{0.2cm}
{\em (a) The proof of learning accuracy:} From the convexity of the loss function, we obtain the following inequality:
\begin{align}
    &\Lc_{j,t}^p (\hat{\thetav}_{j,t}) - \Lc_{j,t}^p (\hat{\thetav}_{j,t+1})\nonumber\\
    &\;\;\;\;\;\leq \left\langle \nabla\Lc_{j,t}^p(\hat{\thetav}_{j,t}),\; \hat{\thetav}_{j,t} - \hat{\thetav}_{j,t+1} \right\rangle \nonumber\\
    &\;\;\;\;\; = \left\langle \frac{1}{\sqrt{ \eta}}\nabla\Lc_{j,t}^{p}(\hat{\thetav}_{j,t}),\; \sqrt{\eta}\left(\hat{\thetav}_{j,t} - \hat{\thetav}_{j,t+1}\right) \right\rangle\nonumber\\
    &\;\;\;\;\; \stackrel{(a)}{\leq} \frac{1}{2 \eta} \left\|\nabla\Lc_{j,t}^p(\hat{\thetav}_{j,t})\right\|^2 + \frac{\eta}{2}\left\|\hat{\thetav}_{j,t} - \hat{\thetav}_{j,t+1}\right\|^2,\label{eq:proof5}
\end{align}where (a) is due to the Fenchel-Young inequality \cite{rockafellar1970convex}. From (\ref{eq:proof0}) and (\ref{eq:proof5}), we can get:
\begin{align}
    &\Lc_{j,t}^{p}(\hat{\thetav}_{j,t}) - \Lc_{j,t}^{p}\left(\thetav_{j}^{\star}\right) \leq  \frac{1}{2\rho} \left(\left\|\hat{\lambdav}_{j,t}\right\|^2-\left\|\hat{\lambdav}_{j,t+1}\right\|^2\right)\nonumber\\
    &+\frac{\rho}{2}\left(\left\|\hat{\gammav}_{j,t}-\thetav_j^{\star}\right\|^2 - \left\|\hat{\gammav}_{j,t+1}-\thetav_j^{\star}\right\|^2\right)\nonumber\\
    &+\frac{\eta}{2}\left(\left\|\hat{\thetav}_{j,t}-\thetav_{j}^{\star}\right\|^{2} - \left\|\hat{\thetav}_{j,t+1}-\thetav_{j}^{\star}\right\|^{2}\right)+\frac{1}{2\eta}\left\|\nabla\Lc_{j,t}^p(\hat{\thetav}_{j,t})\right\|^2.\nonumber
\end{align} Using the above inequality and the telescoping sum, we have:
\begin{align}
 &\sum_{t=1}^{T}\Lc_{j,t}^p (\hat{\thetav}_{j,t}) - \Lc_{j,t}^p \left(\thetav_{j}^{\star}\right) \leq \frac{1}{2\rho} \left(\left\|\hat{\lambdav}_{j,1}\right\|^2-\left\|\hat{\lambdav}_{j,T+1}\right\|^2\right)\nonumber\\
     & \;\;\;\;\;\;\;\;\;\;\;\;\;\;+\frac{\rho}{2}\left(\left\|\hat{\gammav}_{j,1}-\thetav_{j}^{\star}\right\|^2 - \left\|\hat{\gammav}_{j,T+1}-\thetav_{j}^{\star}\right\|^2\right)\nonumber\\
     &\;\;\;\;\;\;\;\;\;\;\;\;\;\;+\frac{1}{2 \eta}\sum_{t=1}^{T}\left\|\nabla\Lc_{j,t}^p(\hat{\thetav}_{j,t})\right\|^2\nonumber\\
     &\stackrel{(a)}{\leq} \frac{\rho}{2}\left\|\hat{\gammav}_{j,1}-\thetav_{j}^{\star}\right\|^2 + \frac{1}{2 \eta}\sum_{t=1}^{T}\left\|\nabla\Lc_{j,t}^p(\hat{\thetav}_{j,t})\right\|^2,\label{eq:final}
\end{align} where (a) is due to the fact $\hat{\lambdav}_{j,1} = \zerov$, $\|\hat{\lambdav}_{j,T+1}\|^2\geq0$, and $\|\hat{\gammav}_{j,T+1}-\thetav_j^{\star}\|^2\geq 0$. Finally, from the assumptions {\bf (a1)} and {\bf (a2)}, we can get:
\begin{align}
    (\ref{eq:final}) \leq \frac{\rho C}{2} + \frac{T G}{2\eta}
\end{align} Setting $\rho=\sqrt{T}$ and $\eta=\sqrt{T}$, we can verify that DOKL (in Algorithm 1) achieves the sublinear regret, which completes the proof of the part (a).

\vspace{0.2cm}
{\em (b) The proof of consensus violation:} We first obtain the following upper-bound:
\begin{align}
    &\Big|\sum_{i\in \Nc_j}\left(\hat{f}_{j,t}(\xv_{j,t}) -\hat{f}_{i,t}(\xv_{j,t})\right) \Big|^2\nonumber\\
    &\;\;\;\;\;\;\;\;\;\; =\Big|\sum_{i\in \Nc_j}\left( \hat{\thetav}_{j,t}^{\trasp}\zv_{p}(\xv_{j,t}) - \hat{\thetav}_{i,t}^{\trasp}\zv_{p}(\xv_{j,t}) \right)\Big|^2\nonumber\\
    &\;\;\;\;\;\;\;\;\;\; \stackrel{(a)}{\leq}\Big\|\sum_{i\in\Nc_{j}} \left(\hat{\thetav}_{j,t}-\hat{\thetav}_{i,t}\right)\Big\|^2\Big\|\zv_{p}(\xv_{j,t})\Big\|^2\nonumber\\
    &\;\;\;\;\;\;\;\;\;\; \stackrel{(b)}{\leq} \Big\|\sum_{i\in\Nc_{j}} \left(\hat{\thetav}_{j,t}-\hat{\thetav}_{i,t}\right)\Big\|^2,\label{eq:regret_d1}
\end{align} where (a) follows the Cauchy-Schwartz inequality and (b) is due to the fact that $\|\zv(\xv_{j}^t)\|_2^2\leq 1$ from (\ref{eq:zv}). Also, the following inequality is obtained:
\begin{align}
    &\left\|\sum_{i\in\Nc_{j}} \left(\hat{\thetav}_{j,t+1}-\hat{\thetav}_{i,t+1}\right)\right\|^2 \nonumber\\
    &\stackrel{(a)}{=} \frac{4}{\rho^2}\left\| \hat{\lambdav}_{j,t+1}-\hat{\lambdav}_{j,t}\right\|^2=4\left\|\hat{\thetav}_{j,t+1}-\hat{\gammav}_{j,t+1}\right\|^2 \nonumber\\
    &\stackrel{(b)}{\leq} 4\left\|\hat{\thetav}_{j,t+1} - \hat{\gammav}_{j,t}\right\|^2\nonumber\\
    &\stackrel{(c)}{\leq} \frac{8B}{\rho} + \frac{4}{\rho^2} \left(\left\|\hat{\lambdav}_{j,t}\right\|^2-\left\|\hat{\lambdav}_{j,t+1}\right\|^2\right)\nonumber\\
    &\;\;\;\;\;+4\left(\left\|-\thetav_{j}^{\star}+\hat{\gammav}_{j,t}\right\|^2 - \left\|-\thetav_{j}^{\star}+\hat{\gammav}_{j,t+1}\right\|^2\right)\nonumber\\
    &\;\;\;\;\;+\frac{4\eta}{\rho}\left(\frac{1}{2}\left\|\hat{\thetav}_{j,t}-\thetav_{j}^{\star}\right\|^{2} - \frac{1}{2}\left\|\hat{\thetav}_{j,t+1}-\thetav_{j}^{\star}\right\|^{2}\right),\label{eq:regret_d2}
\end{align} where (a) is due to the fact that $\hat{\thetav}_{j,t+1}-\hat{\gammav}_{j,t+1}=\frac{1}{\rho}(\hat{\lambdav}_{j,t} - \hat{\lambdav}_{j,t+1})$, (b) follows the \cite[Lemma 3]{wang2013online}, and (c) is from the rearrangement of (\ref{eq:proof0}) in Lemma~\ref{lem2} and using the assumption {\bf (a3)}. From (\ref{eq:regret_d1}) and (\ref{eq:regret_d2}), and using the telescoping sum, we can get:
\begin{align}
{\rm Regret}_{\rm d}^j (T)&=\sum_{t=1}^{T} \Big|\sum_{i\in \Nc_j}\left(\hat{f}_{j,t}(\xv_{j,t}) -\hat{f}_{i,t}(\xv_{j,t})\right) \Big|^2\nonumber\\
    &\leq \sum_{t=1}^{T}\Big\|\sum_{i\in\Nc_{j}} \hat{\thetav}_{j,t+1}-\hat{\thetav}_{i,t+1}\Big\|^2\nonumber\\
    &\leq \frac{8B T}{\rho}+ 4\left\|\hat{\gammav}_{j,1}-\thetav_{j}^{\star}\right\|^2+\frac{2\eta}{\rho}\left\|\hat{\thetav}_{j,1}-\thetav_{j}^{\star}\right\|^2\nonumber\\
    &\leq \frac{8B T}{\rho}+4C + \frac{4\eta}{\rho}C.\label{eq:thm1-3}
\end{align} Setting $\rho=\sqrt{T}$ and $\eta=\sqrt{T}$, the sublinear regret is achieved, which completes the proof.

\subsection{Proof of Theorem 2}\label{subsec:proof2}

We prove the sublinear regrets of DOMKL in Algorithm 2.

\vspace{0.2cm}
{\em (a) The proof of learning accuracy:} 
We first provide the following key lemma:
\vspace{0.1cm}
\begin{lemma}\label{lem3} Setting $\eta_g = \Oc(\sqrt{T})$ and using the weights in (\ref{eq:weight-update}), the following sublinear regret is achieved:
\begin{align}
        &\sum_{t=1}^{T} \Lc\left(\sum_{p=1}^{P}\hat{q}_{j,t}^p\hat{f}_{j,t}^p(\xv_{j,t}),y_{j,t}\right) \nonumber\\
        &\;\;\;\;\;\;\;\;\;\;\;\;\;\;\;\; - \min_{1\leq p \leq P} \sum_{t=1}^T \Lc\Big(f_{p}^{\star}(\xv_{j,t}),y_{j,t}\Big)\leq \Oc(\sqrt{T}).\label{eq:M_proof2}
    \end{align} 
\end{lemma}
\begin{IEEEproof} The proof is provided in Appendix~\ref{app:lem3}.
\end{IEEEproof}
From Theorem~\ref{thm1}, we know that for any kernel $p \in [P]$, 
\begin{align}
        \sum_{t=1}^{T} \Lc_{j,t}^{p}(\hat{\thetav}_{j,t+1}) - \Lc_{j,t}^p(\thetav_{j}^{\star}) \leq \Oc(\sqrt{T}).\label{eq:M_proof1}
\end{align} 
By integrating (\ref{eq:M_proof1}) and (\ref{eq:M_proof2}), the proof is completed.


{\em (b) The proof of consensus violation:}  We first obtain the following upper-bound on the discrepancy at time $t$:
\begin{align}
    &\left|\sum_{i\in\Nc_j} \left( \sum_{p=1}^{P}\hat{q}_{j,t}^p\hat{f}_{j,t}^p(\xv_{j,t}) - \sum_{p=1}^{P}\hat{q}_{i,t}^p\hat{f}_{i,t}^p(\xv_{j,t})\right) \right|^2\nonumber\\
    &\stackrel{(a)}{\leq}2\left|\hat{q}_{j,t}^p\sum_{i\in\Nc_j} \left(\sum_{p=1}^{P}\hat{f}_{j,t}^p(\xv_{j,t}) - \sum_{p=1}^{P}\hat{f}_{i,t}^p(\xv_{j,t})\right) \right|^2+ 2D_t\nonumber\\
    &\leq 2\sum_{p=1}^{P} \left(\hat{q}_{j,t}^p\right)^2 \left|\sum_{i\in\Nc_j} \left(\left(\hat{\thetav}_{j,t}^{p}\right)^{\trasp}\zv_{p}(\xv_{j,t}) -   \left(\hat{\thetav}_{i,t}^{p}\right)^{\trasp}\zv_{p}(\xv_{j,t})\right) \right|^2\nonumber\\
    &\;\;\;\;\; +2D_t\nonumber\\
    &\leq 2\sum_{p=1}^{P} \left(\hat{q}_{j,t}^p\right)^2 \left\|\sum_{i\in\Nc_{j}} \left(\hat{\thetav}_{j,t}^{p}-\hat{\thetav}_{i,t}^{p}\right)\right\|^2+2D_t,\label{eq:proof-thm2-1}
\end{align} where (a) is due to the fact that $(A+B)^2\leq 2(A^2+B^2)$ and 
\begin{align}
D_t = \left|\sum_{i\in \Nc_j} \sum_{p=1}^{P}\Big(\hat{q}_{j,t}^p - \hat{q}_{i,t}^p\Big)\hat{f}_{i,t}^p\Big(\xv_{j,t}\Big) \right|^2.
\end{align}
From (\ref{eq:proof-thm2-1}), we obtain the following upper-bound:
\begin{align}
    &{\rm Regret}_{d}^j(T) \nonumber\\
    &\;\; \leq 2\sum_{p=1}^{P} \left(\hat{q}_{j,p}^t\right)^2 \sum_{t=1}^{T}\left\|\sum_{i\in\Nc_{j}} \left(\hat{\thetav}_{j,t}^{p}-\hat{\thetav}_{i,t}^{p}\right)\right\|^2+2\sum_{t=1}^TD_t\nonumber\\ 
    &\;\; \stackrel{(a)}{\leq} 2\sum_{p=1}^{P} \left(\hat{q}_{j,p}^t\right)^2 \left(\frac{8B T}{\rho}+4C+ \frac{4\eta}{\rho}C\right)+2\sum_{t=1}^TD^t\nonumber\\
    &\;\; \stackrel{(b)}{\leq} \frac{16 B T}{\rho}+8C + \frac{8\eta}{\rho}C+2\sum_{t=1}^TD_t,\label{eq:proof10}
\end{align} where (a) is from (\ref{eq:thm1-3}) and (b) is due to the fact that $\sum_{p=1}^{P}(\hat{q}_{j,t}^p)^2\leq 1$. Also, we have that
\begin{align}
    \sum_{t=1}^T D_t &\leq |\Nc_j|\sum_{p=1}^{P}\left(\hat{q}_{j,t}^p - \hat{q}_{i,t}^p\right)^2\left\|\hat{\thetav}_{j,t}^p\right\|^2\nonumber\\
    &\stackrel{(a)}{\leq} |\Nc_j|C P\sum_{t=1}^{T}\epsilon_t^2\nonumber\\
    &\stackrel{(b)}{\leq} \Oc(\sqrt{T}),\label{eq:proof11}
\end{align} where (a) is from the assumption {\bf (a2)} and (b) is from the assumption {\bf (a4)}. From (\ref{eq:proof10}) and (\ref{eq:proof11}), and setting $\rho=\sqrt{T}$ and $\eta=\sqrt{T}$, the sublinear regret of discrepancy is achieved, which completes the proof.

\begin{figure}[!t]
\centerline{\includegraphics[width=7cm]{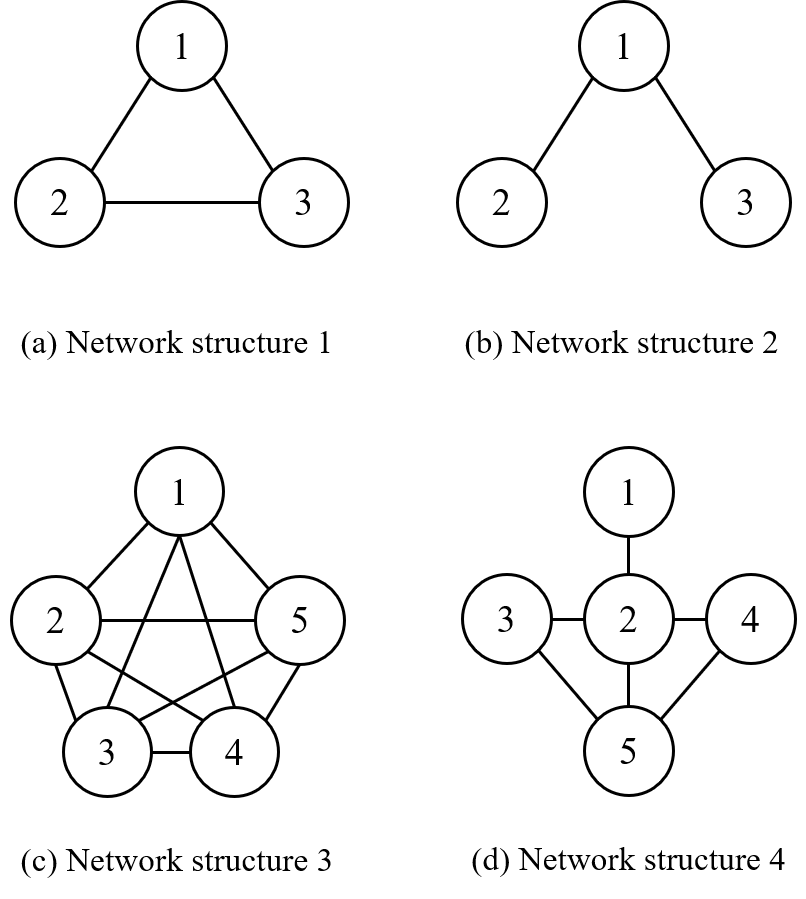}}
\caption{Summary of network structures (learner-connectivity) for experiments.}
\label{fig:NC}
\end{figure}

%
%

%
%
\section{Experiments}\label{sec:Exp}

We provide the experimental tests to verify the effectiveness of the proposed DOMKL. In particular, we consider the online   regression and time-series prediction tasks with various real-world datasets. A regularized least-square loss function (with respect to $\thetav$) is considered, which is defined as
\begin{equation*}
    \Lc\left(\thetav^{\trasp}\zv_{p}(\xv_{t}), y_{t}\right)= \left[y_{t} - \thetav^{\trasp}\zv_{p}(\xv_{t})\right]^2 + \lambda \|\thetav\|^2,
\end{equation*} for some regularization parameter $\lambda>0$.
Regarding the distributed settings, the various network structures (i.e., the various network connectivity among learners) are considered, which are described in Fig.~\ref{fig:NC}. Also, to verify the superiority of the proposed algorithms, we compare the performances with the state-of-the-art centralized counterpart (named online multiple kernel learning (OMKL)) proposed in \cite{shen2019random, hong2020active}. Especially, we will demonstrate the effectiveness of the proposed DOMKL into two-fold: 1) it can achieve the performance of the best kernel function (under the distributed setting) in hindsight; 2) it can achieve the performance of OMKL without sharing local data directly. To accomplish this, the following methods are used in our experiments:

\begin{itemize}
    \item {\bf OMKL}: The {\it centralized} online {\it multiple} kernel learning algorithm using online gradient decent (OGD) method \cite{shen2019random, hong2020active}. 
    \item {\bf DOKL1} : The proposed {\it decentralized} online {\it single} kernel learning in Algorithm 1 having the Gaussian kernel with $\sigma^{2}=10^{-2}$.
    \item {\bf DOKL2} : The proposed {\it decentralized} online {\it single} kernel learning algorithm in Algorithm 1 having the Gaussian kernel with $\sigma^{2}=10^{2}$.
    \item {\bf DOMKL}: The proposed {\it decentralized} online {\it multiple} kernel learning in Algorithm 2, where we use the kernel dictionary consisting of the 17 Gaussian kernels whose parameters are given as
\begin{equation}\label{eq:GK}
    \sigma_{p}^{2}=10^{\frac{i-9}{2}},~ p=1,\dots,17.
\end{equation}
\end{itemize} Additionally, the hyper-parameters of the proposed DOKL and DOMKL are set by
\begin{equation}
    \rho=\eta=\eta_{g}=\sqrt{T} ~\text{and}~ D=50.
\end{equation} Obviously these hyper-parameters can control the tradeoff between learning accuracy and consensus constraint. Unfortunately, the optimization of such hyper-parameters are not straightforward as in the centralized OMKL \cite{shen2019random,hong2020active}, which is left of a future work. Instead, we have chosen the hyper-parameters on the basis of analytical results in Section~\ref{sec:TA}, where it was proved that DOMKL can yield an optimal performance asymptotically  with $\rho=\eta=\eta_{g}=\sqrt{T}$. 

To show the effectiveness of using multiple kernels, DOMKL is compared with the best single kernel, where it is obtained via an exhaustive search by assuming that all data is known in advance. In detail, the exhaustive search among 17 Gaussian kernels in (\ref{eq:GK}) is conducted on Wave energy data to find the best single kernel. This reveals that in this dataset, the Gaussian kernel with $\sigma^{2}=10^{-2}$ (i.e., DOKL1) is the optimal. Accordingly, DOKL shows the best performance for Wave energy data (see Tables II and III). It is remarkable that this process to find the optimal single kernel is not applicable, since at every time $t$, the optimization of a function learning should be performed only with the partial datasets 
$\{(\xv_{\tau},y_{\tau}):\tau\in[t]\}$. Thus, it is not possible to find the best kernel function over the entire dataset. From this comparison, we can confirm that the proposed DOMKL achieves the optimal performance on top of its practical usefulness.

In the following subsections, the performances of DOMKL are investigated into two aspects: {\it node-wise} and {\it network-wise}, which are respectively evaluated by learning accuracy and consensus violation.

{\em i) Learning accuracy:} For every learner $j \in \mathcal{V}$, the accuracy of a function learning is evaluated by the standard mean-square-error (MSE) as
\begin{equation}
\mbox{MSE}= \frac{1}{T}\sum_{t=1}^{T}\left(\hat{y}_{j,t}-y_{j,t}\right)^{2} ~ \;\;\text{for} ~  j \in \mathcal{V}, 
\end{equation} where $\hat{y}_{j,t}=\hat{f}(\xv_{j,t})$ and $y_{j,t}$ denote an estimated label and a true label of the learner $j$ at time $t$, respectively. Clearly, a smaller value of MSE indicates a more precise learned function. However, this metric cannot capture the consensus among neighboring learners, which is evaluated from the metric in the below.



{\em ii) Consensus violation:} The network-wise consensus is evaluated by measuring the differences of the learning functions of neighboring learners. For the consistency with the theoretical analysis in Section~\ref{sec:TA}, we take the following metric:
\begin{equation}
    \mbox{CV}=\frac{1}{T}\sum_{t=1}^{T} \left|\sum_{i\in \Nc_j}\left(\hat{f}_{j,t}(\xv_{j,t}) -\hat{f}_{i,t}(\xv_{j,t})\right) \right|^2.
\end{equation} Clearly, a smaller value of the above metric can ensure that neighboring learners in a network generate a more similar estimated functions. Note that due to the randomness caused by the above algorithms (e.g., $\zv(\xv)$ in \eqref{eq:zv}), the averaged MSE and CV performances over 10 trials are evaluated.





\begin{table}
{\caption{Summary of Real Datasets for Experiments}}
\vspace{-0.2cm}
\label{tb:DataSummary}
\centering
\begin{tabular}{ c||c|c|c } 
\hline
\multicolumn{4}{c}{\cellcolor{lightgray}\bf Regression task} \\ \hline
Datasets & \# of features & \# of data & feature type  \\
\hline
Weather  & 21 & 7750 & real  \\ 
\hline
Conductivity  & 81 & 11000 & real \\ 
\hline
Air quality  & 13 & 7322 & real \\ 
\hline
Wave energy  & 48  & 9500 & real \\ 
\hline
\multicolumn{4}{c}{\cellcolor{lightgray}\bf Time series prediction task} \\ \hline
Datasets & \# of features & \# of data & feature type  \\
\hline
Traffic  & 5,10 & 6500 & real  \\ 
\hline
Temperature  & 5,10 & 5500 & real \\ 
\hline
\end{tabular}
\end{table}


\begin{table}[t]
{\caption{Comparisons of MSE $(\times 10^{-2})$ on online regression tasks}}
\label{tb:MSE_n3}
\vspace{-0.2cm}
\begin{center}
\begin{tabular}{l||c|c|c|c|c}
\hline
         \multicolumn{6}{c}{\cellcolor{lightgray} \bf Network Structure 1} \\ \hline
& &  \multicolumn{1}{l|}{Weather} & 
         \multicolumn{1}{l|}{Conductivity} & 
         \multicolumn{1}{l|}{Air} &
         \multicolumn{1}{l}{Wave} \\ \hline
OMKL &  & 0.133  & 2.03 & 0.22 & 0.028 \\ \hline
\multirow{3}{*}{DOKL1} & Learner 1 & 0.425 & 4.22 & 0.72  & 0.048 \\ 
\cline{2-6} & Learner 2 & 0.389 & 3.63 & 0.93 & 0.047 \\ 
\cline{2-6} & Learner 3 & 0.612 & 4.02 & 0.843 & 0.047\\ \hline
\multirow{3}{*}{DOKL2} & Learner 1 & 9.402  & 9.28 & 1.819 & 0.297 \\ \cline{2-6} & Learner 2 & 8.455 & 9.81 & 2.096  & 0.346 \\ 
\cline{2-6} & Learner 3 & 10.15 & 7.61 &  2.188 &  0.328\\ \hline
\multirow{3}{*}{DOMKL} & 
\cellcolor{LightCyan} Learner 1 & \cellcolor{LightCyan}0.190 & \cellcolor{LightCyan}2.15 & \cellcolor{LightCyan}0.232  &  \cellcolor{LightCyan}0.048 \\ \cline{2-6} & \cellcolor{LightCyan}Learner 2 & \cellcolor{LightCyan}0.176 & \cellcolor{LightCyan}2.11 & \cellcolor{LightCyan}0.241  & \cellcolor{LightCyan}0.047 \\ 
\cline{2-6} & \cellcolor{LightCyan}Learner 3 & \cellcolor{LightCyan}0.247 & \cellcolor{LightCyan}1.68 &  \cellcolor{LightCyan}0.312 & \cellcolor{LightCyan}0.047 \\ \hline
\multicolumn{6}{c}{\cellcolor{lightgray}\bf  Network Structure 2} \\ \hline
& & \multicolumn{1}{l|}{Weather} & 
         \multicolumn{1}{l|}{Conductivity} & 
         \multicolumn{1}{l|}{Air} & 
         \multicolumn{1}{l}{Wave} \\ \hline
OMKL &  & 0.133  & 2.03 & 0.22 & 0.028 \\ \hline
\multirow{3}{*}{DOKL1} & Learner 1 & 0.424 & 4.21 & 0.722 & 0.048 \\ 
\cline{2-6} & Learner 2 & 0.35 & 3.49 &  0.869  & 0.047 \\ 
\cline{2-6} & Learner 3 & 0.56 & 3.80 &   0.772 & 0.047\\ 
\hline
\multirow{3}{*}{DOKL2} & Learner 1 & 8.52 & 8.53 &  1.861 & 0.283  \\ 
\cline{2-6} & Learner 2 & 6.74 & 9.56 & 2.729 & 0.283 \\ 
\cline{2-6} & Learner 3 & 12.37 & 8.17 &  2.20 & 0.361\\ 
\hline
\multirow{3}{*}{DOMKL} & \cellcolor{LightCyan}{Learner 1} & \cellcolor{LightCyan}{0.187} & \cellcolor{LightCyan}{2.11} & \cellcolor{LightCyan}{0.225} & \cellcolor{LightCyan}{0.046} \\ \cline{2-6} &\cellcolor{LightCyan}{Learner 2} & \cellcolor{LightCyan}{0.186} & \cellcolor{LightCyan}{2.15} & \cellcolor{LightCyan}{0.28} & \cellcolor{LightCyan}{0.049} \\ 
\cline{2-6} & \cellcolor{LightCyan}{Learner 3} & \cellcolor{LightCyan}{0.254} & \cellcolor{LightCyan}{1.69} & \cellcolor{LightCyan}{0.344} & \cellcolor{LightCyan}{0.049}  \\ 
\hline
\end{tabular}
\end{center}
\end{table}

\begin{table}
{\caption{Comparisons of MSE $(\times 10^{-2})$ on online regression tasks}}
\label{tb:MSE_n5}
\vspace{-0.2cm}
\begin{center}
\begin{tabular}{l||c|c|c|c|c}
\hline
         \multicolumn{6}{c}{\cellcolor{lightgray} \bf Network Structure 3} \\ \hline
& & \multicolumn{1}{l|}{Weather} & 
         \multicolumn{1}{l|}{Conductivity} &
         \multicolumn{1}{l|}{Air} & 
         \multicolumn{1}{l}{Wave} \\ \hline
OMKL & & 0.133  & 2.03 & 0.22 & 0.028 \\ \hline
\multirow{5}{*}{DOKL1} & Learner 1 & 0.462 & 3.97 & 0.81  & 0.071 \\ \cline{2-6} & Learner 2 & 0.68 & 4.59  & 0.7 & 0.068  \\ \cline{2-6} & Learner 3 & 0.389 & 3.74  & 0.89 & 0.071\\ \cline{2-6} & Learner 4 &0.677 & 3.70 & 1.24 & 0.071\\ \cline{2-6} & Learner 5 & 0.54 & 5.55 & 0.93 & 0.071 \\ \hline
\multirow{5}{*}{DOKL2} & Learner 1 & 11.09 & 10.58 & 2.644 & 0.339  \\ \cline{2-6} & Learner 2 & 10.12 & 11.63  & 1.773 & 0.395 \\ \cline{2-6} & Learner 3 & 8.17 & 11.26  & 2.649 & 0.358 \\ \cline{2-6} & Learner 4 & 10.85 & 9.59 & 3.772 & 0.366 \\ \cline{2-6} & Learner 5 & 7.55 & 9.69 & 2.70 & 0.4 \\ \hline
\multirow{5}{*}{DOMKL} & \cellcolor{LightCyan}{Learner 1} & \cellcolor{LightCyan}{0.262}  & \cellcolor{LightCyan}{2.45} & \cellcolor{LightCyan}{0.352}  & \cellcolor{LightCyan}{0.072} \\ \cline{2-6} 
&\cellcolor{LightCyan}{Learner 2} & \cellcolor{LightCyan}{0.327} & \cellcolor{LightCyan}{2.52} & \cellcolor{LightCyan}{0.26} &  \cellcolor{LightCyan}{0.069} \\ \cline{2-6} & \cellcolor{LightCyan}{Learner 3} & \cellcolor{LightCyan}{0.245} & \cellcolor{LightCyan}{2.34} & \cellcolor{LightCyan}{0.317} & \cellcolor{LightCyan}{0.071}  \\ \cline{2-6} & \cellcolor{LightCyan}{Learner 4} & \cellcolor{LightCyan}{0.331} & \cellcolor{LightCyan}{2.45} &\cellcolor{LightCyan}{0.51} & \cellcolor{LightCyan}{0.072} \\ \cline{2-6} & \cellcolor{LightCyan}{Learner 5} & \cellcolor{LightCyan}{0.268} & \cellcolor{LightCyan}{1.80} & \cellcolor{LightCyan}{0.42} & \cellcolor{LightCyan}{0.071}\\ \hline
\multicolumn{6}{c}{\cellcolor{lightgray}\bf Network Structure 4} \\ \hline
& & \multicolumn{1}{l|}{Weather} & 
         \multicolumn{1}{l|}{Conductivity} &
         \multicolumn{1}{l|}{Air} & 
         \multicolumn{1}{l}{Wave} \\ \hline
OMKL &  & 0.133  & 2.03 & 0.22 & 0.028 \\ \hline
\multirow{5}{*}{DOKL1} & Learner 1 & 0.42 & 3.83 & 0.79 & 0.068 \\ \cline{2-6} & Learner 2 & 0.68 & 4.59 & 0.70 & 0.068  \\ \cline{2-6} & Learner 3 & 0.36 & 4.53 & 0.84 & 0.068\\ \cline{2-6} & Learner 4 & 0.66 & 4.31 & 1.11 & 0.068\\ \cline{2-6} & Learner 5 & 0.54 & 3.93 & 0.91 & 0.07 \\ \hline
\multirow{5}{*}{DOKL2} & Learner 1 & 10.94 & 12.59 & 3.11 & 0.421 \\ \cline{2-6} & Learner 2 & 6.67 & 9.20 & 1.44  & 0.399 \\ \cline{2-6} & Learner 3 & 8.78 & 15.28 & 2.59 & 0.424 \\ \cline{2-6} & Learner 4 & 11.56 & 9.95 & 3.36 & 0.325 \\ \cline{2-6} & Learner 5 & 8.39 & 8.43 & 2.32 & 0.33 \\ \hline
\multirow{5}{*}{DOMKL} & \cellcolor{LightCyan}{Learner 1} & \cellcolor{LightCyan}{0.282} & \cellcolor{LightCyan}{2.64} & \cellcolor{LightCyan}{0.53} & \cellcolor{LightCyan}{0.077} \\ \cline{2-6} 
&\cellcolor{LightCyan}{Learner 2} & \cellcolor{LightCyan}{0.311} & \cellcolor{LightCyan}{2.35} & \cellcolor{LightCyan}{0.25} & \cellcolor{LightCyan}{0.069} \\ \cline{2-6} & \cellcolor{LightCyan}{Learner 3} & \cellcolor{LightCyan}{0.253} & \cellcolor{LightCyan}{2.37} & \cellcolor{LightCyan}{0.37} & \cellcolor{LightCyan}{0.073}  \\ \cline{2-6} & \cellcolor{LightCyan}{Learner 4} & \cellcolor{LightCyan}{0.336} & \cellcolor{LightCyan}{2.41} & \cellcolor{LightCyan}{0.52} & \cellcolor{LightCyan}{0.074} \\ \cline{2-6} & \cellcolor{LightCyan}{Learner 5} & \cellcolor{LightCyan}{0.275} & \cellcolor{LightCyan}{1.67} & \cellcolor{LightCyan}{0.41} & \cellcolor{LightCyan}{0.072}\\ \hline
\end{tabular}
\end{center}
\end{table}

\begin{table}[t]
{\caption{Comparisons of consensus violation (CV) on online regression tasks}}
\label{tb:MSE}
\vspace{-0.2cm}
\begin{center}
\begin{tabular}{l||c|c|c|c|c}
\hline
         \multicolumn{6}{c}{\cellcolor{lightgray}\bf Network Structure 1} \\ \hline
& & \multicolumn{1}{l|}{Weather} & 
         \multicolumn{1}{l|}{Conductivity} & 
         \multicolumn{1}{l|}{Air} &
         \multicolumn{1}{l}{Wave}\\ \hline
\multirow{3}{*}{DOKL1} & Learner 1 & 0.0017 & 0.05  & 0.017 & 0.00014 \\ 
\cline{2-6} & Learner 2 & 0.0018 & 0.049  & 0.017  & 0.00013\\ 
\cline{2-6} & Learner 3 & 0.002  & 0.05 & 0.013 & 0.00013\\ \hline
\multirow{3}{*}{DOKL2} & Learner 1 & 0.072 & 0.067  & 0.029 & 0.0027 \\ \cline{2-6} & Learner 2 & 0.074 & 0.069  & 0.031 & 0.0028 \\ 
\cline{2-6} & Learner 3 & 0.079 & 0.071  & 0.032 & 0.0031\\ \hline
\multirow{3}{*}{DOMKL} & \cellcolor{LightCyan}{Learner 1} & \cellcolor{LightCyan}{0.0021} & \cellcolor{LightCyan}{0.034}  & \cellcolor{LightCyan}{0.0065} & \cellcolor{LightCyan}{0.0001} \\ \cline{2-6} & \cellcolor{LightCyan}{Learner 2} & \cellcolor{LightCyan}{0.0022} & \cellcolor{LightCyan}{0.036}  & \cellcolor{LightCyan}{0.0054} & \cellcolor{LightCyan}{0.0001} \\ 
\cline{2-6} & \cellcolor{LightCyan}{Learner 3} & \cellcolor{LightCyan}{0.0034} & \cellcolor{LightCyan}{0.039}  & \cellcolor{LightCyan}0.0071 & \cellcolor{LightCyan}0.0001 \\ \hline
\multicolumn{6}{c}{\cellcolor{lightgray}\bf Network Structure 2} \\ \hline
& & \multicolumn{1}{l|}{Weather} & 
         \multicolumn{1}{l|}{Conductivity}  & 
         \multicolumn{1}{l|}{Air}& 
         \multicolumn{1}{l}{Wave}\\ \hline
\multirow{3}{*}{DOKL1} & Learner 1 & 0.0023 & 0.028 & 0.018  & 0.0001\\ 
\cline{2-6} & Learner 2 & 0.0023  & 0.027 & 0.018  & 0.0001\\ 
\cline{2-6} & Learner 3 & 0.0025 & 0.027 &  0.014  & 0.0001\\ \hline
\multirow{3}{*}{DOKL2} & Learner 1 & 0.061 & 0.07 & 0.034  & 0.0029 \\ \cline{2-6} & Learner 2 & 0.06 & 0.067   & 0.031 & 0.0029\\ 
\cline{2-6} & Learner 3 & 0.077 & 0.071   & 0.037 & 0.0035\\ \hline
\multirow{3}{*}{DOMKL} & \cellcolor{LightCyan}{Learner 1} & \cellcolor{LightCyan}{0.0022} & \cellcolor{LightCyan}{0.036}  & \cellcolor{LightCyan}{0.0075} & {0.0001} \\ \cline{2-6} & \cellcolor{LightCyan}{Learner 2} & \cellcolor{LightCyan}{0.0022} & \cellcolor{LightCyan}{0.037}  & \cellcolor{LightCyan}{0.0053} & \cellcolor{LightCyan}{0.0001} \\ 
\cline{2-6} & \cellcolor{LightCyan}{Learner 3} & \cellcolor{LightCyan}{0.0032} & \cellcolor{LightCyan}{0.041}  &\cellcolor{LightCyan}{0.0071} & \cellcolor{LightCyan}{0.0001} \\ \hline
\end{tabular}
\end{center}
\end{table}

\begin{table}
{\caption{Comparisons of consensus violation (CV) on online regression tasks}}
\label{tb:MSE}
\vspace{-0.2cm}
\begin{center}
\begin{tabular}{l||c|c|c|c|c}
\hline
         \multicolumn{6}{c}{\cellcolor{lightgray} \bf Network Structure 3} \\ \hline
& & \multicolumn{1}{l|}{Weather} & 
         \multicolumn{1}{l|}{Conductivity} &
         \multicolumn{1}{l|}{Air} & 
         \multicolumn{1}{l}{Wave} \\ \hline
\multirow{5}{*}{DOKL1} & Learner 1 & 0.0011  & 0.019 & 0.011 & 0.0001 \\ \cline{2-6} & Learner 2 & 0.0011 & 0.018 & 0.012 & 0.0001  \\ \cline{2-6} & Learner 3 & 0.0012 & 0.018 & 0.012 & 0.0001\\ \cline{2-6} & Learner 4 & 0.0011  & 0.019 & 0.012 & 0.0001\\ \cline{2-6} & Learner 5 & 0.0012 & 0.017 & 0.012 & 0.0001 \\ \hline
\multirow{5}{*}{DOKL2} & Learner 1 & 0.054 & 0.058 & 0.026 & 0.0021  \\ \cline{2-6} & Learner 2 & 0.054 & 0.065 & 0.022 & 0.0021 \\ \cline{2-6} & Learner 3 & 0.055 & 0.064 & 0.024 & 0.002\\ \cline{2-6} & Learner 4 & 0.057  & 0.06 & 0.036 & 0.002\\ \cline{2-6} & Learner 5 & 0.057 & 0.063 & 0.033 & 0.002 \\ \hline
\multirow{5}{*}{DOMKL} & \cellcolor{LightCyan}{Learner 1} & \cellcolor{LightCyan}{0.0026} & \cellcolor{LightCyan}{0.033} & \cellcolor{LightCyan}{0.0054} & \cellcolor{LightCyan}{0.0001} \\ \cline{2-6} 
&\cellcolor{LightCyan}{Learner 2} & \cellcolor{LightCyan}{0.0025} & \cellcolor{LightCyan}{0.041} & \cellcolor{LightCyan}{0.004}  & \cellcolor{LightCyan}{0.0001} \\ \cline{2-6} & \cellcolor{LightCyan}{Learner 3} & \cellcolor{LightCyan}{0.0013} & \cellcolor{LightCyan}{0.038} &\cellcolor{LightCyan}{0.0044} & \cellcolor{LightCyan}{0.0001} \\ \cline{2-6} & \cellcolor{LightCyan}{Learner 4} & \cellcolor{LightCyan}{0.0016} & \cellcolor{LightCyan}{0.036} & \cellcolor{LightCyan}{0.0081} & \cellcolor{LightCyan}{0.0001} \\ \cline{2-6} & \cellcolor{LightCyan}{Learner 5} & \cellcolor{LightCyan}{0.002} & \cellcolor{LightCyan}{0.047} & \cellcolor{LightCyan}{0.0055} & \cellcolor{LightCyan}{0.0001}\\ \hline
\multicolumn{6}{c}{\cellcolor{lightgray}\bf Network Structure 4} \\ \hline
& & \multicolumn{1}{l|}{Weather} & 
         \multicolumn{1}{l|}{Conductivity} &
         \multicolumn{1}{l|}{Air} & 
         \multicolumn{1}{l}{Wave} \\ \hline
\multirow{5}{*}{DOKL1} & Learner 1 & 0.0018 & 0.05 & 0.014 & 0.0001 \\ \cline{2-6} & Learner 2 & 0.0017 & 0.048 & 0.015 & 0.0001   \\ \cline{2-6} & Learner 3 & 0.0018 & 0.045 & 0.013 & 0.0001\\ \cline{2-6} & Learner 4 & 0.0016 & 0.043 & 0.013 & 0.0001\\ \cline{2-6} & Learner 5 & 0.0017 & 0.039 & 0.011 & 0.0001  \\ \hline
\multirow{5}{*}{DOKL2} & Learner 1 & 0.06 & 0.112 & 0.022 & 0.0021  \\ \cline{2-6} & Learner 2 & 0.064 & 0.154 & 0.025 & 0.0021  \\ \cline{2-6} & Learner 3 & 0.053 & 0.128 & 0.022 & 0.002\\ \cline{2-6} & Learner 4 & 0.059 & 0.107 & 0.028 & 0.0022\\ \cline{2-6} & Learner 5 & 0.058 & 0.114 & 0.023 & 0.0025 \\ \hline
\multirow{5}{*}{DOMKL} & \cellcolor{LightCyan}{Learner 1} & \cellcolor{LightCyan}{0.0027} & \cellcolor{LightCyan}{0.038} &\cellcolor{LightCyan}{0.0066} & \cellcolor{LightCyan}{0.0001} \\ \cline{2-6} 
&\cellcolor{LightCyan}{Learner 2} & \cellcolor{LightCyan}{0.0029} & \cellcolor{LightCyan}{0.045} & \cellcolor{LightCyan}{0.0084} & \cellcolor{LightCyan}{0.0001} \\ \cline{2-6} & \cellcolor{LightCyan}{Learner 3} & \cellcolor{LightCyan}{0.0023} & \cellcolor{LightCyan}{0.036} & \cellcolor{LightCyan}{0.0049} & \cellcolor{LightCyan}{0.0001} \\ \cline{2-6} & \cellcolor{LightCyan}{Learner 4} & \cellcolor{LightCyan}{0.0039}  & \cellcolor{LightCyan}{0.035} & \cellcolor{LightCyan}{0.0101} & \cellcolor{LightCyan}{0.0001}\\ \cline{2-6} & \cellcolor{LightCyan}{Learner 5} & \cellcolor{LightCyan}{0.0018}  & \cellcolor{LightCyan}{0.045} &\cellcolor{LightCyan}{0.0065} &\cellcolor{LightCyan}{0.0001}\\ \hline
\end{tabular}
\end{center}
\end{table}

\subsection{Online Regression Tasks}

For the experiments on online regression tasks, we consider the following real datasets from UCI Machine Learning Repository, which are also summarized in Table I.

\begin{itemize}
    \item {\bf Weather} \cite{D.Cho2020} : The data contains 7750 samples obtained from LDAPS model operated by the Korea Meteorological Administration during 2015-2017, of which the feature in $\xv_{t} \in \RR^{21}$ shows the geographical variables of Seoul. The purpose is to predict the minimum temperatures of next day. 
    \item {\bf Conductivity} \cite{K.Hamidieh2018} : The dataset contains 11000 samples of extracted from superconductors, where each feature in in $\xv_{t} \in \RR^{81}$ represents critical information to construct superconductor such as density and mass of atoms. The goal is to predict the critical temperature to create superconductor.
    \item {\bf Air Quality} \cite{SDeVito2008} : This dataset includes 7322 samples, which features include hourly response from an array of 5 metal oxide chemical sensors embedded in an Air Quality multi-sensor device deployed on the field in a city of Italy. The goal is to predict the concentration of polluting chemicals in the air.
    \item {\bf Wave energy} \cite{M.Neshat2018} : This data contains 9500 samples consisted of positions and absorbed power obtained from wave energy converters (WECs) in four real wave scenarios from the southern coast of Australia. The goal is to predict total power energy of the farm.
\end{itemize} The above data samples are divided into all distributed learners in a network uniformly and randomly. Accordingly, each learner has the $T$ number of local data where $T$ is determined as
\begin{equation*}
    T = \left\lfloor \mbox{the number of samples}/\mbox{the number of learners}\right\rfloor.
\end{equation*}

{\bf Performance evaluation:} For online regression tasks, the MSE (learning accuracy) performances of various methods are provided in Table II and III. 
Recall that the network structures are described in Fig.~\ref{fig:NC}. Table II summarizes the MSE performances of DOKL1, DOKL2, and DOMKL, when network structures 1 and 2 in Fig.~\ref{fig:NC} are considered. Similarly, Table III shows their MSE performances when network structures 3 and 4 in Fig.~\ref{fig:NC}) are considered. We first observe that the learning accuracy of DOMKL is notable for all network structures on every real dataset. In particular, it is shown that the proposed DOMKL generates the 5$\sim$6 times more precise function than DOKL2 in all network structures. This clearly validates the effectiveness of using multiple kernels over a single kernel, since in practice, it is quite challenging to find a proper single kernel. For the comparison with the best single kernel (i.e., DOKL1 in Wave data), the proposed DOMKL can achieve the same performance. Namely, DOMKL can approach the best single kernel function in a practical way. Not surprisingly, DOMKL can outperform DOKL1 and DOKL2 on Weather, Conductivity, and Air data. Furthermore, DOMKL demonstrates the comparable performances with the state-of-the-art centralized OMKL. This result is remarkable since the performance of DOMKL is attained by preserving a privacy, whereas in OMKL, all local data should be shared. This can convince the practical merit of the proposed DOMKL in machine learning tasks arising from IoT systems. Generally, the updated estimates of neighboring learners seem to be fairly shared, leading to a similar learning accuracy. Regarding a network-wise perspective, Table IV and V illustrate the amount of consensus violations of the proposed methods. It is clearly shown that each learner in the proposed DOMKL can generate a learned function quite similar to those of neighboring learners. In fact, the consensus violation can be controlled with the hyper-parameter $\eta$. Namely, we can choose the best $\eta$ according to the required learning accuracy and consensus violation. From our experimental results, we can observe that DOMKL yields the competitive performances as well as attractive consensus violations.

\begin{table}[t]
{\caption{Comparisons of MSE $(\times 10^{-2})$ on time-series prediction tasks}}
\label{tb:MSE_n3_ts}
\vspace{-0.2cm}
\begin{center}
\begin{tabular}{l||c|c|c|c|c}
\hline
\multicolumn{6}{c}{\cellcolor{lightgray}\bf Network Structure 1} \\ \hline
& & \multicolumn{2}{c|}{$p=5$} & \multicolumn{2}{c}{$p=10$}\\ \hline
& & \multicolumn{1}{l|}{Traffic} & \multicolumn{1}{l|}{Temp} & \multicolumn{1}{l|}{Traffic} & \multicolumn{1}{l}{Temp} \\ \hline
OMKL &  & 0.992  & 0.022 & 1.011 & 0.022 \\ \hline
\multirow{3}{*}{DOKL1} & Learner 1 & 1.969 & 0.123 & 1.987 & 0.122 \\ 
\cline{2-6} & Learner 2 & 2.124  & 0.117 & 2.127  & 0.117\\ 
\cline{2-6} & Learner 3 & 2.018  & 0.081 & 2.094  & 0.081 \\ \hline
\multirow{3}{*}{DOKL2} & Learner 1 & 17.01 & 0.086  & 23.79 & 0.105 \\ 
\cline{2-6} & Learner 2 & 24.03 & 0.083 & 25.22  & 0.098\\ 
\cline{2-6} & Learner 3 & 18.19 & 0.082 & 27.47  & 0.092 \\ \hline
\multirow{3}{*}{DOMKL} & \cellcolor{LightCyan}{Learner 1} & \cellcolor{LightCyan}{0.991} & \cellcolor{LightCyan}{0.086} & \cellcolor{LightCyan}{1.097}  & \cellcolor{LightCyan}{0.088}\\ \cline{2-6} &\cellcolor{LightCyan}{Learner 2} & \cellcolor{LightCyan}{1.34} & \cellcolor{LightCyan}{0.069} & \cellcolor{LightCyan}{1.488}  & \cellcolor{LightCyan}{0.07}\\ 
\cline{2-6} &\cellcolor{LightCyan}{Learner 3} & \cellcolor{LightCyan}{1.008} & \cellcolor{LightCyan}{0.062} & \cellcolor{LightCyan}{1.141} & \cellcolor{LightCyan}{0.062} \\ \hline

\multicolumn{6}{c}{\cellcolor{lightgray}\bf Network Structure 2} \\ \hline
& & \multicolumn{2}{c|}{$p=5$} & \multicolumn{2}{c}{$p=10$}\\ \hline
&  & \multicolumn{1}{l|}{Traffic} & \multicolumn{1}{l|}{Temp} & \multicolumn{1}{l|}{Traffic} & \multicolumn{1}{l}{Temp} \\ \hline
OMKL  &  & 0.992  & 0.022 & 1.011 & 0.022 \\ \hline
\multirow{3}{*}{DOKL1} & Learner 1 & 1.968 & 0.123  & 1.987 & 0.122 \\ 
\cline{2-6} & Learner 2 & 1.849 & 0.113 & 1.869  & 0.113\\ 
\cline{2-6} & Learner 3 & 1.569 & 0.079 & 1.632 & 0.078\\ \hline
\multirow{3}{*}{DOKL2} & Learner 1 & 20.68 & 0.094 & 18.39  & 0.089 \\ 
\cline{2-6} & Learner 2 & 25.59  & 0.096 & 26.99 & 0.088\\ 
\cline{2-6} & Learner 3 & 25.96 & 0.096 & 26.54 &  0.087 \\ \hline
\multirow{3}{*}{DOMKL} & \cellcolor{LightCyan}{Learner 1} & \cellcolor{LightCyan}{0.981} & \cellcolor{LightCyan}{0.084} & \cellcolor{LightCyan}{1.152} & \cellcolor{LightCyan}{0.084}\\ \cline{2-6} &\cellcolor{LightCyan}{Learner 2} & \cellcolor{LightCyan}{1.338} & \cellcolor{LightCyan}{0.067} & \cellcolor{LightCyan}{1.425} & \cellcolor{LightCyan}{0.067}\\ 
\cline{2-6} & \cellcolor{LightCyan}{Learner 3} & \cellcolor{LightCyan}{0.973} &\cellcolor{LightCyan}{0.062} & \cellcolor{LightCyan}{1.123} & \cellcolor{LightCyan}{0.062}  \\ \hline
\end{tabular}
\end{center}
\end{table}


\begin{table}[t]
{\caption{Comparisons of consensus violation (CV) on time-series prediction tasks}}
\label{tb:Con_n3_ts}
\vspace{-0.2cm}
\begin{center}
\begin{tabular}{l||c|c|c|c|c}
\hline
\multicolumn{6}{c}{\cellcolor{lightgray}\bf Network Structure 1} \\ \hline
& & \multicolumn{2}{c|}{$p=5$} & \multicolumn{2}{c}{$p=10$}\\ \hline
& & \multicolumn{1}{l|}{Traffic} & \multicolumn{1}{l|}{Temp} & \multicolumn{1}{l|}{Traffic} & \multicolumn{1}{l}{Tempe} \\ \hline
\multirow{3}{*}{DOKL1} & Learner 1 & 0.0914 & 0.0001 & 0.0894   &  0.0001 \\ 
\cline{2-6} & Learner 2 & 0.0903 & 0.0001 & 0.089  &  0.0001\\ 
\cline{2-6} & Learner 3 & 0.102  & 0.0001 & 0.0939  & 0.0001 \\ \hline
\multirow{3}{*}{DOKL2} & Learner 1 & 0.1759 & 0.0047 & 0.1271 & 0.0062\\ 
\cline{2-6} & Learner 2 & 0.1689 & 0.0016 & 0.1245   & 0.0020\\ 
\cline{2-6} & Learner 3 & 0.1504 & 0.0009 & 0.1239 & 0.0015 \\ \hline
\multirow{3}{*}{DOMKL} & \cellcolor{LightCyan}{Learner 1} & \cellcolor{LightCyan}{0.0229} & \cellcolor{LightCyan}{0.0008} & \cellcolor{LightCyan}{0.029}  &  \cellcolor{LightCyan}{0.0007}\\ \cline{2-6} &\cellcolor{LightCyan}{Learner 2} & \cellcolor{LightCyan}{0.0183} & \cellcolor{LightCyan}{0.0001} & \cellcolor{LightCyan}{0.027}  & \cellcolor{LightCyan}{0.0001}\\ 
\cline{2-6} & \cellcolor{LightCyan}{Learner 3} & \cellcolor{LightCyan}{0.0251} & \cellcolor{LightCyan}{0.0001} &  \cellcolor{LightCyan}{0.0257} & \cellcolor{LightCyan}{0.0001}  \\ \hline
\multicolumn{6}{c}{\cellcolor{lightgray}\bf Network Structure 2} \\ \hline
& & \multicolumn{2}{c|}{$p=5$} & \multicolumn{2}{c}{$p=10$}\\ \hline
& & \multicolumn{1}{l|}{Traffic} & \multicolumn{1}{l|}{Temp} & \multicolumn{1}{l|}{Traffic} & \multicolumn{1}{l}{Temp} \\ \hline
\multirow{3}{*}{DOKL1} & Learner 1 & 0.0991 & 0.0002 & 0.0979 & 0.0002  \\ 
\cline{2-6} & Learner 2 & 0.0982  & 0.0002 & 0.0975  & 0.0002\\ 
\cline{2-6} & Learner 3 & 0.1130  & 0.0002 & 0.1041 & 0.0002 \\ \hline
\multirow{3}{*}{DOKL2} & Learner 1 & 0.1574 & 0.0096 & 0.1592 & 0.0064 \\ 
\cline{2-6} & Learner 2 &0.1523  & 0.0017 & 0.1541 & 0.0023\\ 
\cline{2-6} & Learner 3 &0.135 & 0.0012 & 0.1369 & 0.0013 \\ \hline
\multirow{3}{*}{DOMKL} & \cellcolor{LightCyan}{Learner 1} & \cellcolor{LightCyan}{0.0353} & \cellcolor{LightCyan}{0.001} & \cellcolor{LightCyan}{0.0497} & \cellcolor{LightCyan}{0.0011}\\ \cline{2-6} &\cellcolor{LightCyan}{Learner 2} & \cellcolor{LightCyan}{0.0208} & \cellcolor{LightCyan}{0.0001} & \cellcolor{LightCyan}{0.0295} & \cellcolor{LightCyan}{0.0002}\\ 
\cline{2-6} & \cellcolor{LightCyan}{Learner 3} & \cellcolor{LightCyan}{0.0282} & \cellcolor{LightCyan}{0.0001} & \cellcolor{LightCyan}{0.0303} & \cellcolor{LightCyan}{0.0001}   \\ \hline
\end{tabular}
\end{center}
\end{table}


\begin{table}
{\caption{Comparisons of MSE $(\times 10^{-2})$ on time-series prediction tasks}}
\label{tb:MSE_n5_ts}
\vspace{-0.2cm}
\begin{center}
\begin{tabular}{l||c|c|c|c|c}
\hline
         \multicolumn{6}{c}{\cellcolor{lightgray} \bf Network Structure 3} \\ \hline
 &   & \multicolumn{2}{c|}{$p=5$} & \multicolumn{2}{c}{$p=10$}\\ \hline
& & \multicolumn{1}{l|}{Traffic} & 
         \multicolumn{1}{l|}{Temp} &
         \multicolumn{1}{l|}{Traffic} & 
         \multicolumn{1}{l}{Temp} \\ \hline
OMKL &  & 0.992  & 0.022 & 1.011 & 0.022 \\ \hline
\multirow{5}{*}{DOKL1} & Learner 1 & 3.046 & 0.194 & 2.937  & 0.193  \\ \cline{2-6} & Learner 2 & 2.553 & 0.119  & 2.513 & 0.118  \\ \cline{2-6} & Learner 3 & 2.703 & 0.191  & 2.824 & 0.192\\ \cline{2-6} & Learner 4 & 2.914 & 0.135 & 2.979 & 0.134\\ \cline{2-6} & Learner 5 & 3.226 & 0.111 & 3.141 & 0.110  \\ \hline
\multirow{5}{*}{DOKL2} & Learner 1 & 20.99 & 0.180 & 27.15 & 0.184 \\ \cline{2-6} & Learner 2 & 20.87 & 0.163  & 21.88 & 0.179  \\ \cline{2-6} & Learner 3 & 27.04 & 0.164  & 24.03 & 0.155 \\ \cline{2-6} & Learner 4 & 24.12 & 0.151 & 25.66 & 0.17\\ \cline{2-6} & Learner 5 & 28.04 & 0.150 & 28.16 & 0.146\\ \hline
\multirow{5}{*}{DOMKL} & \cellcolor{LightCyan}{Learner 1} & \cellcolor{LightCyan}{1.286}  & \cellcolor{LightCyan}{0.147} & \cellcolor{LightCyan}{1.535}  & \cellcolor{LightCyan}{0.148}  \\ \cline{2-6} 
&\cellcolor{LightCyan}{Learner 2} & \cellcolor{LightCyan}{0.969} & \cellcolor{LightCyan}{0.113} & \cellcolor{LightCyan}{1.148} &  \cellcolor{LightCyan}{0.116} \\ \cline{2-6} & \cellcolor{LightCyan}{Learner 3} & \cellcolor{LightCyan}{1.599} & \cellcolor{LightCyan}{0.12} & \cellcolor{LightCyan}{1.838} & \cellcolor{LightCyan}{0.122} \\ \cline{2-6} & \cellcolor{LightCyan}{Learner 4} & \cellcolor{LightCyan}{1.489} & \cellcolor{LightCyan}{0.103} & \cellcolor{LightCyan}{1.689} & \cellcolor{LightCyan}{0.103}\\ \cline{2-6} & \cellcolor{LightCyan}{Learner 5} & \cellcolor{LightCyan}{1.235} & \cellcolor{LightCyan}{0.103} & \cellcolor{LightCyan}{1.595} & \cellcolor{LightCyan}{0.103}\\ \hline
\multicolumn{6}{c}{\cellcolor{lightgray}\bf Network Structure 4} \\ \hline
& & \multicolumn{2}{c|}{$p=5$} & \multicolumn{2}{c}{$p=10$}\\ \hline
 & & \multicolumn{1}{l|}{Traffic} & 
         \multicolumn{1}{l|}{Temp} &
         \multicolumn{1}{l|}{Traffic} & 
         \multicolumn{1}{l}{Temp} \\ \hline
OMKL&  & 0.992  & 0.022 & 1.011 & 0.022 \\ \hline
\multirow{5}{*}{DOKL1} & Learner 1 & 1.690 & 0.185 & 1.673 & 0.184 \\ \cline{2-6} & Learner 2 & 2.539 & 0.118 & 2.50 & 0.118   \\ \cline{2-6} & Learner 3 & 2.145 & 0.185 & 2.177 & 0.185 \\ \cline{2-6} & Learner 4 & 2.172 & 0.131 & 2.207 & 0.130 \\ \cline{2-6} & Learner 5 &  2.648 & 0.110 & 2.642 & 0.109  \\ \hline

\multirow{5}{*}{DOKL2} & Learner 1 & 26.32 & 0.178 & 27.73 & 0.196 \\ \cline{2-6} & Learner 2 & 17.82 & 0.154 & 17.96 & 0.162 \\ \cline{2-6} & Learner 3 & 25.89 & 0.161 & 24.74 & 0.16 \\ \cline{2-6} & Learner 4 & 27.43 & 0.154 & 25.22 & 0.154 \\ \cline{2-6} & Learner 5 &  22.02 & 0.15 & 27.24 & 0.141   \\ \hline

\multirow{5}{*}{DOMKL} &\cellcolor{LightCyan} {Learner 1} & \cellcolor{LightCyan}{1.279} & \cellcolor{LightCyan}{0.150} & \cellcolor{LightCyan}{1.373} & \cellcolor{LightCyan}{0.153}\\ \cline{2-6} & \cellcolor{LightCyan}{Learner 2} & \cellcolor{LightCyan}{0.995} & \cellcolor{LightCyan}{0.111} & \cellcolor{LightCyan}{1.16} & \cellcolor{LightCyan}{0.112} \\ \cline{2-6} & \cellcolor{LightCyan}{Learner 3} & \cellcolor{LightCyan}{1.524} & \cellcolor{LightCyan}{0.114} &\cellcolor{LightCyan}{1.70} &\cellcolor{LightCyan}{0.115} \\ \cline{2-6} & \cellcolor{LightCyan}{Learner 4} &\cellcolor{LightCyan}{1.455} & \cellcolor{LightCyan}{0.101} & \cellcolor{LightCyan}{1.639} & \cellcolor{LightCyan}{0.10} \\ \cline{2-6} & \cellcolor{LightCyan}{Learner 5} & \cellcolor{LightCyan}{1.266} & \cellcolor{LightCyan}{0.104} & \cellcolor{LightCyan}{1.469} & \cellcolor{LightCyan}{0.103} \\ \hline
\end{tabular}
\end{center}
\end{table}


\begin{table}
{\caption{Comparisons of consensus violation (CV) on time-series prediction tasks}}
\label{tb:Con_n5_ts}
\vspace{-0.2cm}
\begin{center}
\begin{tabular}{l||c|c|c|c|c} \hline
         \multicolumn{6}{c}{\cellcolor{lightgray} \bf Network Structure 3} \\ \hline
   &      & \multicolumn{2}{c|}{$p=5$} & \multicolumn{2}{c}{$p=10$}\\ \hline
&  & \multicolumn{1}{l|}{Traffic} & 
         \multicolumn{1}{l|}{Temp} &
         \multicolumn{1}{l|}{Traffic} & 
         \multicolumn{1}{l}{Temp} \\ \hline
\multirow{5}{*}{DOKL1} & Learner 1 & 0.0586 & 0.0001 & 0.0601  & 0.0001 \\ \cline{2-6} & Learner 2 & 0.0604 & 0.0001  & 0.0613 &  0.0001  \\ \cline{2-6} & Learner 3 & 0.0702 & 0.0001  & 0.0734  &  0.0001\\ \cline{2-6} & Learner 4 & 0.0769 & 0.0001 & 0.0798 & 0.0001 \\ \cline{2-6} & Learner 5 & 0.0708 &  0.0001 & 0.0662 & 0.0001 \\ \hline        

\multirow{5}{*}{DOKL2} & Learner 1 & 0.0786 & 0.0026 & 0.0687 & 0.0032\\ \cline{2-6} & Learner 2 &  0.075 & 0.0022 & 0.0681 &  0.0018  \\ \cline{2-6} & Learner 3 & 0.0997  & 0.0037  & 0.0692 &  0.0021\\ \cline{2-6} & Learner 4 & 0.0963 &0.0027  & 0.0723 & 0.0022 \\ \cline{2-6} & Learner 5 & 0.0952 & 0.0015 & 0.0689 & 0.0014 \\ \hline  

\multirow{5}{*}{DOMKL} & 
\cellcolor{LightCyan}{Learner 1} & \cellcolor{LightCyan}{0.0182}  & \cellcolor{LightCyan}{0.0001} & \cellcolor{LightCyan}{0.0002}  &\cellcolor{LightCyan}{0.0001} \\ \cline{2-6} & \cellcolor{LightCyan}{Learner 2} & \cellcolor{LightCyan}{0.0162}  & \cellcolor{LightCyan}{0.0001}  & \cellcolor{LightCyan}{0.0256}  & \cellcolor{LightCyan}{0.0001}\\ \cline{2-6} & \cellcolor{LightCyan}{Learner 3} & \cellcolor{LightCyan}{0.0182} & \cellcolor{LightCyan}{0.0001} & \cellcolor{LightCyan}{0.0302} & \cellcolor{LightCyan}{0.0001} \\ \cline{2-6} & \cellcolor{LightCyan}{Learner 4} & \cellcolor{LightCyan}{0.0189} & \cellcolor{LightCyan}{0.0001}  & \cellcolor{LightCyan}{0.032} & \cellcolor{LightCyan}{0.0001} \\ \cline{2-6} & \cellcolor{LightCyan}{Learner 5} & \cellcolor{LightCyan}{0.0187} & \cellcolor{LightCyan}{0.0001} & \cellcolor{LightCyan}{0.0265} & \cellcolor{LightCyan}{0.0001}\\ \hline

\multicolumn{6}{c}{\cellcolor{lightgray}\bf Network Structure 4} \\ \hline
& & \multicolumn{2}{c|}{$p=5$} & \multicolumn{2}{c}{$p=10$}\\ \hline
&  & \multicolumn{1}{l|}{Traffic} & 
         \multicolumn{1}{l|}{Temp} &
         \multicolumn{1}{l|}{Traffic} & 
         \multicolumn{1}{l}{Temp} \\ \hline
\multirow{5}{*}{DOKL1} & Learner 1 & 0.0859 & 0.0002 & 0.0843 & 0.0002 \\ \cline{2-6} & Learner 2 & 0.0885 & 0.0002 & 0.0863 & 0.0002  \\ \cline{2-6} & Learner 3 & 0.0774 & 0.0001 & 0.0776 &  0.0001\\ \cline{2-6} & Learner 4 & 0.0716 & 0.0001 & 0.0774 & 0.0001 \\ \cline{2-6} & Learner 5 & 0.0625 & 0.0001 & 0.0611 & 0.0001  \\ \hline             

\multirow{5}{*}{DOKL2} & Learner 1 & 0.0979  & 0.0021 & 0.0914 & 0.0021\\ \cline{2-6} & Learner 2 & 0.1014 & 0.0024 & 0.09 & 0.0026  \\ \cline{2-6} & Learner 3 & 0.1023 & 0.0021 & 0.095  & 0.0022 \\ \cline{2-6} & Learner 4 & 0.0965 & 0.0014 & 0.0859 & 0.0016 \\ \cline{2-6} & Learner 5 & 0.1009 & 0.0014 & 0.0902 &  0.0014 \\ \hline      

\multirow{5}{*}{DOMKL} &  \cellcolor{LightCyan}Learner 1 & \cellcolor{LightCyan}{0.0228} & \cellcolor{LightCyan}{0.0002} & \cellcolor{LightCyan}{0.0309} & \cellcolor{LightCyan}{0.0002} \\ \cline{2-6} & \cellcolor{LightCyan}{Learner 2} & \cellcolor{LightCyan}{0.0424} & \cellcolor{LightCyan}{0.0001} & \cellcolor{LightCyan}{0.0502} & \cellcolor{LightCyan}{0.0001}\\ \cline{2-6} & \cellcolor{LightCyan}{Learner 3} & \cellcolor{LightCyan}{0.024} & \cellcolor{LightCyan}{0.0001} & \cellcolor{LightCyan}{0.0311} & \cellcolor{LightCyan}{0.0002} \\ \cline{2-6} & \cellcolor{LightCyan}{Learner 4} & \cellcolor{LightCyan}{0.0229} & \cellcolor{LightCyan}{0.0001} & \cellcolor{LightCyan}{0.0313} & \cellcolor{LightCyan}{0.0001}\\ \cline{2-6} & \cellcolor{LightCyan}{Learner 5} & \cellcolor{LightCyan}{0.0204} & \cellcolor{LightCyan}{0.0001} &\cellcolor{LightCyan}{0.0277} & \cellcolor{LightCyan}{0.0001} \\ \hline
\end{tabular}
\end{center}
\end{table}

\subsection{Time-series Prediction Tasks}

The proposed DOMKL can be naturally extended into time-series prediction tasks which predict the future values in an online distributed fashion. Toward this, we take the popular time-series prediction method called 
Autoregressive (AR) model \cite{mills1991time}. An AR($p$) model predicts the future value $y_{t}$ with the assumption of the linear dependency on its past $p$ values, which is mathematically represented as
\begin{equation}
    y_{t} = c+\sum_{i=1}^{p}\alpha_{i}y_{t-i}+\epsilon_{t},
\end{equation} where $c$ is a constant, $\alpha_{i}$ denotes the weight associated with
$y_{t-i}$, and $\epsilon_{t}$ denotes a Gaussian noise at time $t$. Based on this, the RF-based kernelized AR($p$) model, which can explore a nonlinear dependency, has been introduced in \cite{hong2020active}, where it is formulated as
\begin{align}
y_{t}&=c+f(y_{t-1},y_{t-2},\dots,y_{t-p})+\epsilon_{t}\nonumber\\
&=c+f(\xv_{t})+\epsilon_{t}\nonumber\\
&=c+\thetav^{\trasp}\zv(\xv_{t})+\epsilon_{t},\label{eq:ARp}
\end{align} where $\xv_{t}\triangleq \left[y_{t-1},y_{t-2},\dots,y_{t-p}\right]^{\trasp}$ and $f(\xv_{t})$ belongs to kernel space, which is well-approximated as in \eqref{eq:RFA}. Then, this can be directly plugged into DOMKL framework to solve time-series prediction tasks. The proposed algorithms are tested with the following univariate time-series datasets from UCI Machine Learning Repository:
\begin{itemize}
    \item {\bf Traffic \cite{Timeseries}:} This dataset contains $T$ = 6500 time-series traffic data obtained from Minneapolis Department of Transportation in US. Data is collected from hourly interstate 94 Westbound traffic volume for MN DoT ATR station 301, roughly midway between Minneapolis and St Paul, MN.
    \item {\bf Temperature \cite{Timeseries}:} This dataset consists $T$ = 5500 time-series temperature data obtained from Minneapolis Department of Transportation in US. Data is collected from hourly interstate 94 Westbound temperature for MN DoT ATR station 301, roughly midway between Minneapolis and St Paul, MN.
\end{itemize}

\vspace{0.2cm}
{\bf Performance evaluation:} The learning accuracy (i.e., MSE performance) of various algorithms are provided in Table VI and Table VIII. We consider the network structures in Fig.~\ref{fig:NC} and the AR($p$) model in (\ref{eq:ARp}) with the parameters $p = 5$ and $10$. In all scenarios, the performances of DOMKL are outstanding compared with those of DOKL1 and DOKL2. Especially on traffic data, it proves the 15$\sim$20 times more precise learning ability in every network structure. Even in the comparison with the optimal performances of the best single kernel (e.g., DOKL1), DOMKL proves its remarkable gain. This certifies that the benefit of using multiple kernels in DOMKL actually contributes on the performance improvements, by enlarging a function class. As in online regression tasks, the proposed DOMKL also achieves the almost same performance with OMKL on time-series prediction tasks. This ensures its advantages on data privacy and practicability. From the perspective of network-wise performances, the low values of consensus violation of DOMKL are notable. From Table VII and IX, it is well-convinced that each learner in DOMKL shares the common optimization parameters and less likely violates the consensus constraint, compared with DOKL algorithms.

%
%
\section{Concluding Remarks}\label{sec:con}

In this paper, we proposed a novel distributed online learning framework with multiple kernels. The proposed method is referred to as DOMKL. It was devised by appropriately incorporating an online alternating direction method of multipliers (OADMM) and a distributed Hedge algorithm. To be specific, the former is to optimize kernel functions and the latter is to update the weights for combining the kernel functions collaboratively with neighboring learners in a distributed fashion. The key advantages of the proposed DOMKL are the {\em scalability} (with respect to the number of incoming data) and {\em privacy-preserving}. These merits enable it to be applicable for various function learning tasks arising from IoT systems. Furthermore, we theoretically proved that  DOMKL can achieve an optimal sublinear regret, i.e., it yields the same order with a centralized counterpart (called online multiple kernel learning (OMKL)). This shows that distributed learners (e.g., IoT devices) can achieve the almost same accuracy of the centralized OMKL, without sharing local data directly. 
Beyond the asymptotic analysis, we demonstrated the effectiveness of the proposed DOMKL on online regression and time-series prediction tasks via experimental results with real datasets.

An important future work is to extend the proposed DOMKL into wireless distributed settings, in which the amount of information to be transmitted should be carefully designed according to communication constraints (e.g., channel capacity). Active learning could be a promising solution because in this case, unnecessary information cannot be transmitted by activating useful learners only at every time. Thus, it can reduce the communication cost. Furthermore, this active learning approach can reduce the labeling cost in that each learner only queries some useful incoming data to an oracle. Another promising solution is to employ a quantized OADMM having dynamic quantization levels which are chosen according to the qualities of wireless channels.

\appendices

\section{Proof of Lemma~\ref{lem1-OADMM}}\label{app:lem1}

For ease of exposition we let $|\Vc|=J$. As shown in (\ref{eq:example_A}), from the constructions of $\Am$ and $\Bm$, we have that
\begin{equation}
    \Am\thetav + \Bm\gammav =
    \left[
    \begin{matrix}
   \cv_1\\
   \cv_2\\
   \vdots\\
   \cv_{J}
    \end{matrix}
    \right] \mbox{ with }  \cv_{j} = \left[ \begin{matrix} 
    \thetav_j - \gammav_{\{j,j_1\}}\\
    \vdots\\
    \thetav_j - \gammav_{\{j,j_{|\Nc_j|}\}}\\
    \end{matrix}
    \right],
    \label{eq:const-24}
\end{equation} 
where $j_k$, $k \in \Nc_j$, denotes the indices of the neighboring nodes of node $j$. 
Also, the $\lambdav_{(i,j)}$ denotes the components of $\lambdav$ corresponding to the element $\thetav_{i} - \gammav_{\{i,j\}}$ in $\Am\thetav+\Bm\gammav$. Accordingly, $\lambdav_{(j,i)}$ corresponds to $\thetav_{j} - \gammav_{\{i,j\}}$. Note $(i,j)$ is ordered pair while $\{i,j\}$ is unorderd pair. From (\ref{eq:const-24}) and using the above notations, the optimization of $\thetav_j$ is only associated with the following form:
\begin{align}
    &\hat{\thetav}_{j,t+1} =\argmin_{\thetav_j} \Lc\left(\thetav_j^{\trasp}\zv(\xv_{j,t}), y_{j,t}\right) + \hat{\lambdav}_{j,t}^{\trasp}\thetav_j \nonumber\\
    &\;\;\;\;\;\;\;\;\;\;\;+\frac{\rho}{2}\sum_{i\in\Nc_j}\left\|\thetav_j - \hat{\gammav}_{\{j,i\},t}\right\|^2 + \frac{\eta}{2}\left\|\thetav_j - \hat{\thetav}_{j,t}\right\|^2,\label{eq:thetav}
\end{align} where $\hat{\lambdav}_{j,t} = \sum_{i \in \Nc_j} \hat{\lambdav}_{(j,i),t}$.
Likewise, the optimization problem in (\ref{eq:zv}) can be decomposed as follows:
{
\begin{align}
    &\hat{\gammav}_{\{j,i\},t+1} \nonumber\\
    &= \argmin_{\gammav_{\{j,i\}}}\;\; \hat{\lambdav}_{j,t}^{\trasp}\left(\hat{\thetav}_{j,t+1}-\gammav_{\{j,i\}}\right)+ \hat{\lambdav}_{i,t}^{\trasp}\left(\hat{\thetav}_{i,t+1}-\gammav_{\{j,i\}}\right)\nonumber\\
    &\;\;\;\;\; +\frac{\rho}{2}\left(\left\|\hat{\thetav}_{j,t+1}-\gammav_{\{j,i\}}\right\|^2+\left\|\hat{\thetav}_{i,t+1}-\gammav_{\{j,i\}}\right\|^2\right),
\end{align}}
for $\{j,i\}\in \Ec$. The optimal solution of the above problem can be easily derived as
\begin{align}
&\hat{\gammav}_{\{j,i\},t+1}\nonumber\\ &\;\;\;\;\; =\frac{1}{2}\left(\hat{\thetav}_{j,t+1} + \hat{\thetav}_{i,t+1}\right) + \frac{1}{2\rho}\left(\hat{\lambdav}_{(j,i),t}+\hat{\lambdav}_{(i,j),t}\right).\label{eq:u_z}
\end{align} Also, we can obtain that
\begin{align}
    &\hat{\lambdav}_{(i,j),t+1} = \hat{\lambdav}_{(i,j),t} + \rho\left(\hat{\thetav}_{i,t+1}- \hat{\gammav}_{\{j,i\},t+1}\right)\label{eq:o_lambda}\\
    &\;\;\;\;\;\;\;\; = \frac{\rho}{2}\left(\hat{\thetav}_{i,t+1}-\hat{\thetav}_{j,t+1}\right) + \frac{1}{2}\left(\hat{\lambdav}_{(i,j),t} - \hat{\lambdav}_{(j,i),t}\right). \label{eq:u_lambda}
\end{align} From (\ref{eq:u_lambda}), we observe that the following property holds:
\begin{equation}
    \hat{\lambdav}_{(i,j),t} + \hat{\lambdav}_{(j,i),t} = \zerov. \label{eq:prop}
\end{equation} From (\ref{eq:u_z}) and (\ref{eq:prop}), we have:
\begin{align}
\hat{\gammav}_{\{j,i\},t+1} &=\frac{1}{2}\left(\hat{\thetav}_{j,t+1} + \hat{\thetav}_{i,t+1}\right).\label{eq:zv_2}
\end{align} We can obtain (\ref{eq:theta-update}) by combining 
 (\ref{eq:zv_2}) and (\ref{eq:thetav}). In addition, from (\ref{eq:o_lambda}) and (\ref{eq:zv_2}), we have:
\begin{equation}
     \hat{\lambdav}_{(j,i),t+1} = \hat{\lambdav}_{(j,i),t} + \frac{\rho}{2} \left(\hat{\thetav}_{j,t+1} - \hat{\thetav}_{i,t+1}\right).
\end{equation} Thus, we can get:
\begin{align*}
    \hat{\lambdav}_{j,t+1}&=\sum_{i\in\Nc_j}\hat{\lambdav}_{(j,i),t+1}\\
   &=\hat{\lambdav}_{j,t} + \frac{\rho}{2}\sum_{i\in\Nc_j}\left(\hat{\thetav}_{j,t+1} - \hat{\thetav}_{i,t+1}\right).
\end{align*}
This completes the proof.

\section{Proof of Lemma~\ref{lem2}}\label{app:lem2}

From (\ref{eq:theta-update}), we obtain the following upper-bound on the gradient of the loss function:
\begin{align}
    &\nabla\Lc_{j,t}^p (\hat{\thetav}_{j,t+1}) \nonumber\\
    &= - \Big(\hat{\lambdav}_{j,t+1} + \frac{\rho}{2}\Big(\sum_{i \in \Nc_j}(\hat{\thetav}_{j,t+1}+\hat{\thetav}_{i,t+1}) - \sum_{i\in\Nc_j}(\hat{\thetav}_{j,t} + \hat{\thetav}_{i,t})\Big)\nonumber\\
    &\;\;\;\;\;\;\; +\eta(\hat{\thetav}_{j,t+1} - \hat{\thetav}_{j,t})\Big)\nonumber\\
    &= - \left(\hat{\lambdav}_{j,t+1} + \rho\left(\hat{\gammav}_{j,t+1} - \hat{\gammav}_{j,t}\right) + \eta \left(\hat{\thetav}_{j,t+1} - \hat{\thetav}_{j,t}\right)\right),\label{eq:def_grad}
\end{align} where the last equality follows the definition of $\hat{\gammav}_{j,t}$. From the convexity of the loss function, we obtain:
\begin{align}
    &\Lc_{j,t}^p (\hat{\thetav}_{j,t+1}) - \Lc_{j,t}^p \left(\thetav^{\star}\right)\leq \left\langle \nabla\Lc_{j,t}^p (\hat{\thetav}_{j,t+1}), \hat{\thetav}_{j,t+1} - \thetav^{\star}\right\rangle\nonumber\\
    &\stackrel{(a)}{=} - \left\langle\hat{\lambdav}_{j,t+1}, \hat{\thetav}_{j,t+1}-\thetav^{\star} \right\rangle +\rho\left\langle-\hat{\gammav}_{j,t+1}+\hat{\gammav}_{j,t}, \hat{\thetav}_{j,t+1} - \thetav^{\star}\right\rangle\nonumber\\
    &\;\;\;\;\; -\eta\left\langle\hat{\thetav}_{j,t+1} - \hat{\thetav}_{j,t}, \hat{\thetav}_{j,t+1}-\thetav^{\star} \right\rangle,\label{eq:proof1}
\end{align} where (a) follows (\ref{eq:def_grad}). Using the fact that 
\begin{align*}
    &\left<\vv_1 - \vv_2, \vv_3 + \vv_4\right>\\
    &=\frac{1}{2}\Big(\|\vv_4-\vv_2\|^2-\|\vv_4-\vv_1\|^2 +\|\vv_3+\vv_1\|^2-\|\vv_3+\vv_2\|^2\Big),
\end{align*} we obtain the following inequality:
\begin{align}
    &\left\langle-\hat{\gammav}_{j,t+1}+\hat{\gammav}_{j,t}, \hat{\thetav}_{j,t+1} - \thetav_{j}^{\star}\right\rangle\nonumber \\
    &\;\;\;\;\; = \frac{1}{2}\Big(\left\|-\thetav_{j}^{\star}+\hat{\gammav}_{j,t}\right\|^2 -\left\|-\thetav_{j}^{\star}+\hat{\gammav}_{j,t+1}\right\|^2\nonumber\\
    &\;\;\;\;\;\;\;\;\;\;\;\;\; +\left\|\hat{\thetav}_{j,t+1}-\hat{\gammav}_{j,t+1}\right\|^2-\left\|\hat{\thetav}_{j,t+1} - \hat{\gammav}_{j,t}\right\|^2\Big).\label{eq:proof2}
\end{align} Also, the following inequality  holds:
\begin{align}
    &- \left\langle\hat{\lambdav}_{j,t+1}, \hat{\thetav}_{j,t+1}-\thetav^{\star} \right\rangle+\frac{\rho}{2} \left\|\hat{\thetav}_{j,t+1}-\hat{\gammav}_{j,t+1}\right\|^2 \nonumber\\
    &\stackrel{(a)}{\leq}- \left\langle\hat{\lambdav}_{j,t+1}, \hat{\thetav}_{j,t+1}-\thetav^{\star} \right\rangle - \left\langle\hat{\lambdav}_{j,t+1}, \thetav^{\star} - \hat{\gammav}_{j,t+1} \right\rangle\nonumber\\
    &\;\;\;\;\; +\frac{\rho}{2} \left\|\hat{\thetav}_{j,t+1}-\hat{\gammav}_{j,t+1}\right\|^2,\nonumber\\
    &= - \left\langle\hat{\lambdav}_{j,t+1}, \hat{\thetav}_{j,t+1}-\hat{\gammav}_{j,t+1} \right\rangle+\frac{\rho}{2} \left\|\hat{\thetav}_{j,t+1}-\hat{\gammav}_{j,t+1}\right\|^2,\nonumber\\
    &\stackrel{(b)}{=}  - \frac{1}{\rho}\left\langle\hat{\lambdav}_{j,t+1}, \hat{\lambdav}_{j,t} - \lambdav_{j,t+1}\right\rangle + \frac{1}{2\rho}\left\|\hat{\lambdav}_{j,t} - \lambdav_{j,t+1}\right\|^2\nonumber\\
    &=\frac{1}{2\rho}\left(\left\|\hat{\lambdav}_{j,t}\right\|^2-\left\|\hat{\lambdav}_{j,t+1}\right\|^2\right),\label{eq:proof3}
\end{align}where (a) follows the fact that $\left<\hat{\lambdav}_{j,t+1}, \thetav^{\star} - \hat{\gammav}_{j,t+1}\right> \leq 0$ (see the proof of \cite[Lemma 2]{wang2013online}) and (b) is due to the fact that
\begin{equation*}
    \hat{\thetav}_{j,t+1}-\hat{\gammav}_{j,t+1} =\frac{1}{\rho}\left( \hat{\lambdav}_{j,t} - \hat{\lambdav}_{j,t+1}\right).
\end{equation*} Finally, we have:
\begin{align}
    &-\left\langle\hat{\thetav}_{j,t+1} - \hat{\thetav}_{j,t}, \hat{\thetav}_{j,t+1}-\thetav^{\star} \right\rangle\nonumber\\
    &=\frac{1}{2}\left(\left\|\hat{\thetav}_{j,t}-\thetav^{\star}\right\|^{2} - \left\|\hat{\thetav}_{j,t+1}\thetav^{\star}\right\|^{2} -\left\|\hat{\thetav}_{j,t+1}-\hat{\thetav}_{j,t}\right\|^2\right).\label{eq:proof4}
\end{align}
By integrating (\ref{eq:proof1}), (\ref{eq:proof2}), (\ref{eq:proof3}), and (\ref{eq:proof4}), we complete the proof.

\section{Proof of Lemma~\ref{lem3}}\label{app:lem3}

Recall that the weights for multiple kernels are determined such as
\begin{equation}
    \hat{q}_{j,t+1}^{p} = \frac{\hat{w}_{j,t+1}^{p}\cdot \prod_{i\in\Nc_j}\hat{w}_{i,t+1}^{p}}{\sum_{i=1}^P \Big(\hat{w}_{j,t+1}^{p}\cdot \prod_{i\in\Nc_j}\hat{w}_{i,t+1}^{p}\Big)},
\end{equation} for some parameter $\eta_g>0$, where
\begin{equation}
    \hat{w}_{j,t+1}^{p} = \exp\left(-\frac{1}{\eta_g} \sum_{\tau=1}^t \Lc\left(\hat{f}_{j,\tau}^{p}(\xv_{j,\tau}), y_{j,\tau}\right) \right).
\end{equation} To simplify the notation, we let $\Jc\eqdef\{j\}\cup\{\Nc_j\}$.
For any fixed $j$, we define:
\begin{align}
    \zeta& \eqdef \sum_{t=1}^{T}\log\left(\sum_{p=1}^{P}\hat{q}_{j,t}^{p}\exp\left(-\frac{1}{\eta_g}\sum_{i\in \Jc}\Lc\Big(\hat{f}_{i,t}^p(\xv_{i,t}),y_{i,t}\Big)\right)\right)\nonumber\\
    &\stackrel{(a)}{=}\sum_{t=1}^{T}\EE\left[\exp\left(-\frac{1}{\eta_g}\sum_{i\in\Jc}\Lc\Big(\hat{f}_{i,t}^{I}(\xv_{i,t}),y_{i,t}\Big)\right)\right]\nonumber\\
    &\stackrel{(b)}{\leq} \sum_{i\in \Jc}\left(\sum_{t=1}^{T} -\frac{1}{\eta_g}\EE\Big[\Lc\Big(\hat{f}_{i,t}^{I}(\xv_{i,t}),y_{i,t}\Big)\Big]\right) + \frac{|\Nc_{j}|T L_u^2}{8\eta_g^2},\label{lem3-1}
\end{align} where (a) is due to the fact that $I$ denotes a random variable with the probability mass function (PMF) $(\hat{q}_{j,t}^1,...,\hat{q}_{j,t}^P)$ and (b) follows the Hoeffding inequality with the bounded random variable $\Lc(\hat{f}_{i,t}^{I}(\xv_{i,t}),y_{i,t})$ \cite{wainwright2019high}. Letting $\hat{W}_{t}^{p} \eqdef \hat{w}_{j,t}^{p}\cdot \prod_{i\in\Nc_j}\hat{w}_{i,t}^{p}$, we derive the lower bound on $\zeta$ as follows:
\begin{align}
    \zeta &= \sum_{t=1}^{T}\log\left(\sum_{p=1}^{P}\hat{q}_{j,t}^{p}\exp\left(-\frac{1}{\eta_g}\sum_{i\in \Jc}\Lc\Big(\hat{f}_{i,t}^p(\xv_{i,t}),y_{i,t}\Big)\right)\right)\nonumber\\
    &\stackrel{(a)}{=}\sum_{t=1}^{T}\log\left(\frac{\sum_{p=1}^{P}\hat{W}_{t+1}^{p}}{\sum_{p=1}^{P}\hat{W}_{t}^p}\right)\nonumber\\
    &\stackrel{(b)}{=}\log\left(\sum_{p=1}^{P}\hat{W}_{T+1}^{p}\right) - \log\left(\sum_{p=1}^{P}\hat{W}_{1}^{p}\right)\nonumber\\
    &\leq \log\left(\sum_{p=1}^{P}\hat{W}_{T+1}^{p}\right) - \log{P}\nonumber\\
    &\leq -\frac{1}{\eta_g}\sum_{i\in\Jc} \left(\min_{1\leq p\leq P}\sum_{t=1}^{T} \Lc\Big(\hat{f}_{i,t}^{p}(\xv_{i,t}),y_{i,t}\Big)\right) - \log{P}, \label{lem3-2}
\end{align} where (a) follows the definition of $\hat{q}_{j,t}^p$ and $\hat{W}_{t}^p$ and (b) is due to the telescoping sum. From (\ref{lem3-1}) and (\ref{lem3-2}), we have:
\begin{align*}
    -\frac{1}{\eta_g}\sum_{i\in\Jc} \left(\min_{1\leq p\leq P}\sum_{t=1}^{T} \Lc\Big(\hat{f}_{i,t}^{p}(\xv_{i,t}),y_{i,t}\Big)\right) - \log{P}\\
    \leq \sum_{i\in \Jc}\left(\sum_{t=1}^{T} -\frac{1}{\eta_g}\EE\Big[\Lc\Big(\hat{f}_{i,t}^{I}(\xv_{i,t}),y_{i,t}\Big)\Big]\right) + \frac{|\Nc_j|T L_u^2}{8\eta_g^2}.
\end{align*} By rearranging the above inequality, we can get:
\begin{align*}
    &\sum_{t=1}^{T}\EE\left[\Lc\left(\hat{f}_{j,t}^{I}(\xv_{j,t}),y_{j,t}\right)\right] - \min_{1\leq p\leq P}\sum_{t=1}^{T} \Lc\left(\hat{f}_{j,t}^{p}(\xv_{j,t}),y_{j,t}\right)\\
    &\leq \sum_{i\in\Nc_j}\left[\min_{1\leq p\leq P} \sum_{t=1}^{T}\Lc\left(\hat{f}_{i,t}^{p}(\xv_{i,t}),y_{i,t}\right)\right.\\
    &\;\;\;\;\;\;\;\;\;\;\;\left.-\sum_{t=1}^{T} \EE\left[\Lc\left(\hat{f}_{i,t}^{I}(\xv_{i,t}),y_{i,t}\right)\right] \right]  + \frac{ T L_u^2}{8\eta_g} + \eta_g\log{P}\\
    &\stackrel{(a)}{\leq} \frac{ T L_u^2}{8\eta_g} + \eta_g\log{P},
\end{align*} where (a) follows the fact that for any $i \in \Vc$,
\begin{equation*}
   \min_{ 1\leq p \leq P} \sum_{t=1}^T \Lc\left(\hat{f}_{i,t}^{p}(\xv_{i,t}),y_{i,t}\right)\leq\sum_{t=1}^{T}\EE\left[\Lc\left(\hat{f}_{i,t}^{I}(\xv_{i,t}),y_{i,t}\right)\right].
\end{equation*} Finally, setting $\eta_g = \sqrt{T}$, the sublinear regret is achieved, which completes the proof.

\section*{Acknowledgment}
This work was supported by the National Research Foundation of Korea(NRF) grant funded by the Korea government(MSIT) (NRF-2020R1A2C1099836).


\ifCLASSOPTIONcaptionsoff
  \newpage
\fi



\end{document}

%% file: macros.tex
\long\def\comment#1{}

\newfont{\bbb}{msbm10 scaled 700}

\newfont{\bb}{msbm10 scaled 1100}

\newcommand{\RR}{\mbox{\bb R}}

\newcommand{\EE}{\mbox{\bb E}}

\newcommand{\cv}{{\bf c}}

\newcommand{\pv}{{\bf p}}

\newcommand{\vv}{{\bf v}}
\newcommand{\xv}{{\bf x}}

\newcommand{\zv}{{\bf z}}
\newcommand{\zerov}{{\bf 0}}

\newcommand{\Am}{{\bf A}}
\newcommand{\Bm}{{\bf B}}

\newcommand{\Id}{{\bf I}}

\newcommand{\Ec}{{\cal E}}

\newcommand{\Gc}{{\cal G}}
\newcommand{\Hc}{{\cal H}}

\newcommand{\Jc}{{\cal J}}

\newcommand{\Lc}{{\cal L}}

\newcommand{\Nc}{{\cal N}}
\newcommand{\Oc}{{\cal O}}

\newcommand{\Vc}{{\cal V}}
\newcommand{\Xc}{{\cal X}}
\newcommand{\Yc}{{\cal Y}}

\newcommand{\gammav}{\hbox{\boldmath$\gamma$}}

\newcommand{\lambdav}{\hbox{\boldmath$\lambda$}}

\newcommand{\thetav}{\hbox{\boldmath$\theta$}}

\newcommand{\eqdef}{\stackrel{\Delta}{=}}

\newcommand{\trasp}{{\sf T}}
\newcommand{\transp}{{\sf T}}
